\pdfoutput=1

\documentclass[11pt,a4paper]{article}
\usepackage[hyperref]{emnlp2020}
\usepackage{times}
\usepackage{amsfonts}
\usepackage[export]{adjustbox}
\usepackage{multirow}
\usepackage{latexsym}
\usepackage{graphicx}
\usepackage{subcaption}
\usepackage{amssymb}
\usepackage{mathtools}
\usepackage{amsmath,bm, amsthm}
\usepackage{comment}
\usepackage{url}
\usepackage{dsfont}
\usepackage{stmaryrd}
\usepackage{mathrsfs}
\usepackage{microtype}

\usepackage{latexsym}

\aclfinalcopy 

\newtheorem*{remark}{Remark}

\title{Target Model Agnostic Adversarial Attacks with Query Budgets on Language Understanding Models}

\author{Jatin Chauhan, Karan Bhukar, Manohar Kaul \\
	\texttt{\{chauhanjatin100,choudhary121,manohar.kaul\}@gmail.com} \\
}
\date{}

\DeclareMathOperator*{\argmax}{arg\,max}
\DeclareMathOperator*{\argmin}{arg\,min}

\hfuzz=\maxdimen

\newcount\hbadness
\newdimen\hfuzz

\begin{document}
\maketitle
\begin{abstract}
Despite significant improvements in natural language understanding models with the advent of models like BERT and XLNet, these neural-network based classifiers are vulnerable to blackbox adversarial attacks, where the attacker is only allowed to query the target model outputs. We add two more realistic restrictions on the attack methods, namely limiting the number of queries allowed (query budget) and crafting attacks that easily transfer across different pre-trained models (transferability), which render previous attack models impractical and ineffective. Here, we propose a  target model agnostic adversarial attack method with a high degree of attack transferability across the attacked models. 
Our empirical studies show that in comparison to baseline methods, our method generates highly transferable adversarial sentences under the restriction of limited query budgets.
\end{abstract}

\section{Introduction}
Recent advancements in language understanding models have significantly pushed forth the accuracy achieved on various challenging NLP benchmarks. Despite this success, deep learning models in general are shown to misclassify when small, often humanly imperceptible, perturbations are added to original samples. These perturbed samples are referred to as \emph{adversarial samples} and such an attack is termed an \emph{adversarial attack}. 
The positive side of such adversarial attacks is that apart from exposing vulnerabilities, they also aid improved understanding and interpretation of models by uncovering training biases. 
Various analyses have been performed via crafted adversarial sentences: model sensitivities to small perturbations~\citep{DBLP:journals/corr/LiMJ16a}, testing neural machine translation~\citep{belinkov2018synthetic}, evaluating reading comprehension~\citep{DBLP:journals/corr/JiaL17} and classification models, to name a few.

Adversarial samples can either be generated in a \emph{whitebox} setting, where the target model's parameters like computed gradients, weights, etc., are fully accessible to the attacker or in a \emph{blackbox} setting, where the attacker can only access the inputs and outputs of the target model~\cite{papernot2016,Papernot2017}. For several commercial or proprietary pre-trained models deployed online, only a blackbox setting can be considered as a realistic setting.  
\begin{table}[tbp]
        \small		
		\centering
		\setlength\tabcolsep{7pt}
		\begin{tabular}{p{1.1cm}p{3.8cm}p{1.1cm}}
		 \\
        \hline
        \textbf{}  & \multicolumn{1}{c}{\bf Sentence} & \textbf{Label} \\
        \hline
        \multicolumn{3}{c}{\bf \emph{SST-2}}\\
        \hline
        Original & This is so \textcolor{blue}{bad}. & \textbf{Negative} \\
        Adversarial & This is so \textcolor{red}{wicked}. & \textbf{Positive} \\
        \hline
        &\\
        \multicolumn{3}{c}{\bf \emph{QNLI}}\\
        \hline
        Premise & Who was the game's leading rusher? & \\
        Original & Anderson was the game's leading \textcolor{blue}{rusher} with 90 yards and a touchdown, along with four \textcolor{blue}{receptions} for 10 yards. & \multirow{4}{0.7cm}{\bf \centering Entailment} \\
        Adversarial & Anderson was the game's leading \textcolor{red}{shredder} with 90 yards and a touchdown, along with four \textcolor{red}{receipts} for 10 yards. & \multirow{4}{0.7cm}{\bf \centering Not Entailment} \\
        \hline
		\end{tabular}
		\caption{Target-agnostic adversarial examples generated by our method inducing errors for all targeted models, where each original sentence was predicted correctly.}
		\label{table:sample_adv}
\end{table}

Existing blackbox attacks craft adversarial samples by repeatedly querying (in the order of thousands) a target model to pick those samples that achieve the greatest drop in accuracy~\citep{DBLP:journals/corr/abs-1907-11932,zhang-etal-2019}. This also means that the adversarial sample generation is \emph{specific} to a single pre-trained model and does not generalize well to attacks on other models.

Motivated by the aforementioned observations, we consider added restrictions on the traditional blackbox model to characterize restrictions in real-world systems. These restrictions render all previous blackbox attacks impractical or infeasible. We introduce the following limited settings:\\
\textbf{Query-limited setting}: Here, the service that deploys a trained model imposes an upper limit on the number of queries to the model due to resource constraints. Upon exceeding such a limit, additional monetary costs can be levied per query.\\
\textbf{Transferability setting}: Here an attack method, rather than constructing adversarial samples specific to a target model, should instead generate adversarial samples that can be used to attack a diverse range of pre-trained target models. In other words, the attack method should possess \emph{high attack transferability} across distinct target models.\\
Our proposed attack works well under both the added restrictions by \emph{not querying} target models during training, unlike previous methods. Rather, we construct a set of adversarial candidate samples \emph{just once} in an off-line fashion by first identifying \emph{important words} in a sentence using an enhanced neural language model On-LSTM~\citep{shen2018ordered} with MultiHead attention and replacing them with specifically chosen synonyms in such a manner that the resulting adversarial sentence maintains a very similar distribution to the input data upon which the target models are trained. Furthermore, we impose upper bounds on: (i) the number of important words that can be replaced per sentence and (ii) the number of adversarial sentences generated per original sentence. Table~\ref{table:sample_adv} shows an example of adversarial sentences produced by our attack method.\\
\noindent\textbf{Contributions}: 
(i) \emph{Target-agnostic:} We propose a \emph{target model agnostic} attack on language understanding models that works under a limited query budget. 
(ii) \emph{Transferability:} We employ offline training to generate a set of adversarial sentences \emph{just once}, which can then be efficiently used in \emph{repeated attacks} against several pre-trained models across a wide range of NLP tasks.
(iii) \emph{Empirical studies:} We conducted an exhaustive empirical analysis to gain deeper insights into our target-agnostic attack and we achieve substantial drops in accuracy of the target models. 
For example, our methods drops the accuracy of \emph{BERT} on \emph{MNLI} by $21\%$ (from $82.7\%$ to $61.63\%$) with an extremely tight budget of $20$, which is more than $15$ times the accuracy drops by the baseline attack methods. We point that our method considers a weaker black-box setting in which we have some data available to train the ON-LSTM model. This can be the same data over which the attacked NLP model is trained, or it can have similar underlying distribution. However, the amount of data used to train ON-LSTM in our experiments is substantially low (less than 10K samples per dataset).

\section{Related Work}
Adversarial attacks have gained widespread popularity in computer vision~\citep{Goodfellow2014ExplainingAH, DBLP:journals/corr/abs-1802-08195, DBLP:journals/corr/abs-1711-01991, Xie_2017_ICCV}, as they are \emph{continuous input 
domains} and therefore lend easily to attacks based on gradient searches and randomization. 
However, NLP tasks present a new modality, i.e., \emph{discrete input domains} on which extending 
gradient based attack methods are not straightforward.
There have been several attempts to perform gradient based attacks via GANs to create natural language adversarial samples~\citep{zhao2018generating}, however the generated adversarial samples are of an inferior quality.
Other works apply \emph{heuristic strategies} to generate adversarial samples by first identifying important features of the text (which could be characters, words or even sentences) followed by the application of greedy search strategies to perturb these features, while obeying constraints that preserve the quality of generated text. Most existing whitebox~\citep{ebrahimi-hotflip, DBLP:journals/corr/SamantaM17} as well as blackbox~\citep{iyyer-adversarial, DBLP:journals/corr/abs-1907-11932, zhao2018generating} methods have used such search heuristics to successfully generate adversarial samples by maintaining the semantic and syntactic properties of the original text. Recent works~\citep{Behjati2019UniversalAA, wallace2019universal} have focused on generating universal tokens to attack different models with good generalization properties, while~\citep{ribeiro-semantically, iyyer-adversarial} tried to create adversaries via paraphrasing. We direct the reader to~\citep{Zhang2019GeneratingTA} for a detailed survey of adversarial attack methods.

\section{Target Model Agnostic Attacks}
\begin{figure*}[tbp]
		\centering
		\begin{subfigure}{.55\textwidth}
			\includegraphics[width=70mm, height=40mm]{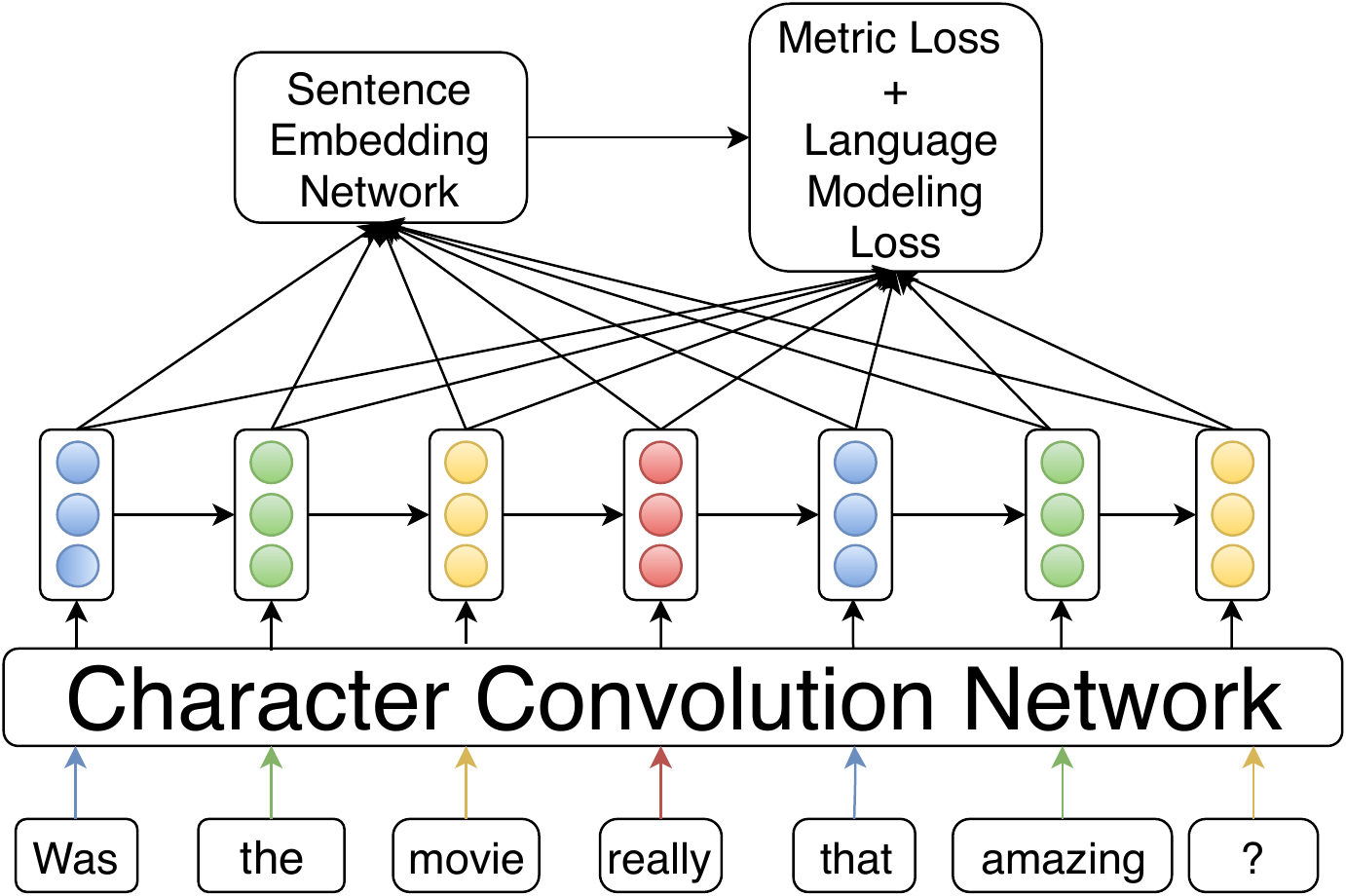}
			\caption{Model}
		\end{subfigure}%
		\begin{subfigure}{.3\textwidth}
			\centering
			\includegraphics[width=40mm, height=40mm]{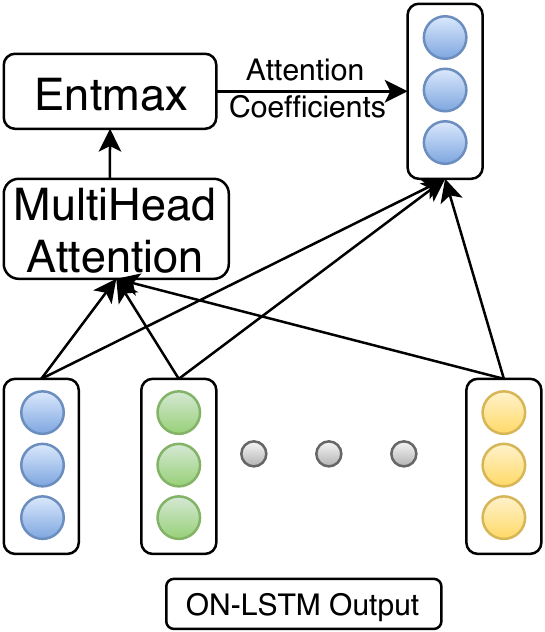}
			\caption{Sentence Embedding Network}
		\end{subfigure}%
		\caption{Our On-LSTM network architecture enhanced with sparse attention via \texttt{entmax}.}
		\label{fig:Class_acc_diffx}
\end{figure*}

\subsection{Problem definition}
\label{prob_formulation}
We consider a pre-trained target model $f:\mathcal{X} \rightarrow \mathcal{Y}$ trained on a set $\mathcal{X}$ of i.i.d. observations, each associated with a class label from set $\mathcal{Y}$. 
For a classification task, the class label belongs to a finite set of predefined discrete values, whereas for a regression task the class label is a \emph{continuous real-valued scalar}.

For the pre-trained model $f$ and a normal sample $\bm{x}$ (sentence)\footnote{We use the term \emph{sample} and \emph{sentence} interchangeably} with class label $y$, 
the goal of an \emph{untargeted} adversarial attack is to find an adversarial sample $\bm{x_{adv}}$
that is nearly indistinguishable from $\bm{x}$, with respect to human perception, such that $f(\bm{x_{adv}}) \neq y$ (i.e., $\bm{x_{adv}}$ gets misclassified to an arbitrary class other than $y$). 
In a black-box setting, attackers are only allowed access to the final outputs of $f$. 
Typical black-box attacks consist of repeatedly querying (in the order of thousands) a target model $f$ with a sample $\bm{x'}$ to greedily improve the sample's quality until a satisfactory $\bm{x_{adv}}$ is crafted, akin to a \emph{denial of service (DoS) attack} on $f$. Such attacks are both cumbersome to train and completely impractical when attacking trained models deployed online.  

Instead, we propose an untargeted blackbox attack, where a set of \emph{candidate adversarial samples} $\mathcal{X}_{c}$ are generated during training \emph{without querying} the target model $f$ and the cardinality of $\mathcal{X}_{c}$ does not exceed a \emph{budget} $K$. 
We define $\mathcal{X}_{c}$ formally as:
\[
\mathcal{X}_{c} = \lbrace \bm{x'} \mid
\bm{x} \in \mathcal{X} , \mathcal{S}(\bm{x},\bm{x'}) \geq \epsilon, |\mathcal{X}_{c}| \leq K   \rbrace
\]
where $\mathcal{S}(\cdot,\cdot)$ is a \emph{semantic similarity} measure between sentences. In words, 
$\mathcal{X}_{c}$ contains at most $K$ adversarial candidates 
whose semantic similarity to sample $\bm{x}$ exceeds a user-defined threshold $\epsilon$.
We are interested in finding the \emph{largest subset of true adversarial samples} $X_{adv}$ among the set of adversarial candidates $\mathcal{X}_{c}$.
\begin{remark}
	We decouple the necessity to query a specific target model $f$ from adversarial sample generation and therefore arrive at a \emph{target-agnostic} attack method where $\mathcal{X}_{c}$ is generated once and is subsequently used to repeatedly attack a variety of models trained on $\mathcal{X}$.
	\end{remark}

\subsection{Our target-agnostic attack method}
In this section, we describe our proposed target-agnostic attack method that does not generate adversarial samples specific to any given target model.
Our approach consists of two steps:(i) identification of \emph{important words} in a sentence via a neural language attention model, and (ii) replacement of these important words to create adversarial samples conforming to the requirements of preserving the semantic as well as syntactic structure of the original sample enhanced via task-specific language modeling to get adversarial samples following the distribution of original samples.

\subsubsection{Identifying important words}
Our method uses the recent state-of-the-art language model called 
\emph{On-LSTM}~\citep{shen2018ordered}, which induces a \emph{tree-structured hierarchy} on the hidden states of the LSTM network via \emph{monotonic master input} and \emph{master forget} gates. This inductive bias allows the On-LSTM model to perform tree-like composition operations, obtaining noteworthy improvements on tasks like language modeling and unsupervised constituency parsing.
We enhance the On-LSTM model with a \emph{multi-head attention} mechanism to identify important words upon which classification is performed via a \emph{classifier MLP} alongside the standard language modeling objective. Figure~\ref{fig:Class_acc_diffx} shows the architecture of our enhanced On-LSTM model.

Given a sequence of tokens $S=(x_1,\cdots,x_N)$, we generate a corresponding sequence of 
word embeddings $S_{emb}=(\bm{e_1},\cdots,\bm{e_N})$.
To handle out-of-vocabulary (OOV) words as well as to perform character level attacks, we use \emph{Character-level CNNs}~\citep{DBLP:journals/corr/ZhangZL15} to construct word embeddings. 
Given a token $x_i$, we perform 1-D convolutions over its character embeddings stored in the lookup 
matrix $\bm{F} \in \mathds{R}^{|C| \times d_c}$, where $|C|$ is the vocabulary size of characters and 
$d_c$ is the character embedding size, using 3 different kernels of sizes in $\{3, 4, 5\}$ and $100$ output 
channels, which are concatenated and passed through a linear layer to obtain the final word embedding 
$\bm{e_i}$. Repeating this process for every token, the new sequence $S_{emb}$ thus obtained serves as the input to the first layer of our network. We use 
a $2$-layer On-LSTM network with the standard update procedures as described 
in~\citep{shen2018ordered}.

Let $\bm{H}$ = $[\bm{h_1}, \cdots, \bm{h_N}]$ be the matrix consisting of the output vectors from the 
final layer of the model. We pass these through a \texttt{MultiHead} attention block as described 
in~\citep{DBLP:journals/corr/VaswaniSPUJGKP17}. The output is then passed through a 2-layer MLP  with a SeLU~\citep{DBLP:journals/corr/KlambauerUMH17} activation unit, which we call $C^{Att}$ to 
obtain scalar attention coefficients for each output vector $h_i$ as: 
\begin{align}
\bm{P}  &= \texttt{MultiHead}(H) \\ \nonumber
\bm{P'} &= C^{Att}(P)
\end{align}
where $\bm{P'} = [p_1, ....p_N]$ consists of scalar attention values corresponding to each output vector of the matrix $\bm{H}$. These attention values are then normalized via the \texttt{entmax}~\citep{peters-etal-2019-sparse} normalization function:
\begin{equation}
\label{eq:entamx_att}
\alpha = \texttt{entmax}( \bm{P'} )
\end{equation}
Intuitively, each of the output vectors $\bm{h_i}$ contains rich context information about the local n-gram chunks of the sentence. The \texttt{MultiHead} attention block allows chunk-to-chunk attention and the updated embeddings in $\bm{P}$ contain context relative information throughout the sequence, aiding the LSTM in capturing dependencies throughout the sequence. We specifically use \texttt{entmax} as our normalization function as the sparsity induced by it allows the model to learn to focus on the most important words of the sentence. Thus, serving as our \textit{important word identification unit}. The final task classification is performed by passing the combination of the vectors $\bm{h_i}$ weighted by $\alpha$ using a 3-layer MLP which we call $C^{clf}$. More formally, 
\begin{equation}
y = \texttt{softmax} (C^{clf}(\bm{H}\alpha^T))
\end{equation}
In tasks with paired-sentences, we perform 3 operations: element wise addition, subtraction and multiplication, followed by concatenation of the corresponding embedding vectors $\bm{H}\alpha^T$ of the sentence pairs before passing them through the classifier $C^{clf}$.
Our final cross-entropy loss related objective function is:
\begin{equation}
J(\theta) = \sum_{i=1}^{m}t_{i}log(y_i) + L(\bm{H}, S) + \lambda \lVert \theta \rVert^2
\end{equation}
where $m$ is the number of classes, $t \in \mathbb{R}^m$ is the one-hot representation of the ground truth, $y \in \mathbb{R}^m$ is the estimated class probability, $L$ is the standard language modeling objective over the output vectors of the model, and $\lambda$ is the L2 regularization hyper-parameter. In a regression task, the cross-entropy loss is replaced by \textit{mean-squared error (MSE)} loss. 


The language modeling objective function allows our attack method to learn a task-specific sentence distribution that has both \emph{low perplexity} and \emph{low divergence} of the crafted adversarial sentences with respect to the input data distribution. 
After training the model, we use the attention coefficients ($\alpha$), to identify the \emph{important words} in the sentence during adversarial generation phase as described in the following section.

\subsubsection{Generating adversarial sentences}
\label{adv_generation_section}
Having described the procedure for identifying important words in a sentence, we now describe the procedure to generate adversarial sentences. We limit both the \emph{maximum number of adversarial samples} ($K$) as well as the \emph{maximum number of words} ($M$) that can be perturbed in the original sentence to generate an adversarial sentence. Our objective is to generate a set of, at most $K$, \emph{distinct} adversarial samples $\mathcal{X}_{c}$ for a given normal sample $\bm{x}$, which can then cause maximum misclassification in every target model. 
We consider the following steps to generate $\mathcal{X}_{c}$:\\
(a) Consider a single token $x$ in a sentence and its corresponding word embedding $\bm{e}$. 
Note that these word embeddings are specifically curated for synonym extraction~\citep{mrksic-etal-2016-counter}.
We then compute a set $E_x$ of the $k$-most-semantically-similar words to $x$, 
based on the cosine-similarity between $\bm{e}$ and the embeddings of neighboring words.
The set $E_x$ is further reduced by pruning: (i) neighboring words  whose cosine similarity with the original token $x$'s embedding falls below a user-defined threshold $\phi$, (ii) neighbors whose part-of-speech (POS) tags do not match the POS tag of $x$, and (iii) stop-words. We denote this reduced set of words by $E'_x$. Such a set is computed per token.\\
(b) Recall that each word is associated with an \emph{attention coefficient} $\alpha$ (as described in Equation~\ref{eq:entamx_att}). In a given sentence, we pick a set $X$ of top-$M$ words according to their corresponding $\alpha$ values. Now, we must replace each word $\bm{x} \in X$ with a semantically-similar word from set $E'_x$. The total number of possible substitutions is upper bounded by $M^{|E^{'}_x|}$, which is too large and therefore we randomly sample $R$ combinations and accordingly replace the original words with their corresponding semantically-similar words to arrive at our final adversarial candidate sentences. In our experiments, $R$ is set to $600$.\\
(c) Each candidate sentence in $R$ is assigned a \emph{perplexity score} when passed through our On-LSTM model. On the basis of this score, we pick the top-$W$ candidates with the lowest perplexity scores to further reduce the size of $R$. For each of the $W$ candidates, we then compute their \emph{sentence semantic similarity} w.r.t to the original sentence using the \emph{Infersent} model~\citep{DBLP:journals/corr/ConneauKSBB17}. Finally, we retain the top-$K$ candidates ranked by decreasing order of their \emph{sentence semantic similarity} score and further prune away candidates whose score is below $\epsilon$ (as defined in Section~\ref{prob_formulation}). The successful candidates from this set are referred to as $X_{adv}$, defined in Section~\ref{prob_formulation}. In our experiments, $K$ is set to $20$, as explained in section \ref{analysis_sec}.

\begin{table*}[tbp]
\tiny

\centering
\setlength\tabcolsep{3.5pt}
\begin{tabular}{lrrrrrrrrrrrrrr}
\\
\hline
\textbf{}   &  \multicolumn{5}{c}{\bf \emph{BERT}}  &  \multicolumn{5}{c}{\bf \emph{XLNeT}} & \multicolumn{4}{c}{\bf \emph{BiLSTM + Attn + ELMo}} \\
\cline{2-5}   \cline{7-10} \cline{12-15}
&  SST-2  & MNLI(m / mm) & QQP  & QNLI &  &  SST-2  & MNLI(m / mm) & QQP  & QNLI &  &  SST-2  & MNLI(m / mm) & QQP  & QNLI \\

\hline
\emph{Pre-attack Accuracy} & 92.14 & 82.7 / 84.88 & 90.5 & 84.65 &  & 93.3 & 85.00 / 85.38 & 88.93 & 83.21 &  & 88.68 & 70.1 / 71.36 & 81.4 & 76.2 \\                                                      
&\\

\hline
\textbf{} &  \multicolumn{14}{c}{\bf \emph{Method: TextFooler}} \\
\hline
\emph{Avg. Accuracy Drop} & \underline{0.90} & \underline{1.21} / \underline{1.18} & 1.04 & \underline{0.83} & & \underline{0.78} & \underline{1.54} / \underline{1.44} & 1.19 & \underline{1.03} &  & \underline{1.32} & \underline{1.98} / 1.87 & 1.05 & \underline{0.88} \\
\emph{Max. Accuracy Drop} & 5.51 & 26.58 / 29.56 \textbf{**} & 23.68\textbf{**} & 6.35 &  & 5.30\textbf{**} & 26.90 / 27.40 \textbf{**} & 23.53\textbf{**} & 2.71 &  & 7.14 & 26.43 / 28.38 \textbf{**} & 23.28\textbf{*} & 7.49\textbf{**} \\
\emph{Semantic Similarity} & 0.83 & 0.81 / 0.79 & 0.78 & 0.87 & & 0.82 & 0.73 / 0.75 & 0.79 & 0.82 &  & 0.83 & 0.72 / 0.73 & 0.78 & 0.86 \\

&\\

\hline
\textbf{} &  \multicolumn{14}{c}{\bf \emph{Method: Genetic}} \\
\hline
\emph{Avg. Accuracy Drop} & 0.73 & 1.11 /1.09 & \underline{1.23} & 0.69 &  & 0.75 & 1.28 / 1.31 & \underline{1.49} & 0.88 &  & 1.27 & 1.83 / \underline{1.97} & \underline{1.31} & 0.79 \\
\emph{Max. Accuracy Drop} & 7.45\textbf{**} & 23.79 / 25.31 & 22.78 & 7.02\textbf{**} &  & 3.66 & 22.75 / 22.56 & 21.00 & 5.98\textbf{**} &  & 8.55\textbf{**} & 24.21 / 26.40 & 17.34 & 7.09 \\
\emph{Semantic Similarity} & 0.77 & 0.72 / 0.70 & 0.76 & 0.81 & & 0.81 & 0.67 / 0.69 & 0.71 & 0.79 &  & 0.76 & 0.64 / 0.62 & 0.72 & 0.80 \\

&\\
\hline
\textbf{} &  \multicolumn{14}{c}{\bf \emph{Method: Ours}} \\
\hline
\emph{Avg. Accuracy Drop} & \textbf{5.97} & \textbf{21.07} / \textbf{20.31} & \textbf{6.84} & \textbf{5.01} &  & \textbf{5.81} & \textbf{21.96} / \textbf{21.48} & \textbf{6.20} & \textbf{4.59} &  & \textbf{7.84} & \textbf{23.47} / \textbf{22.93} & \textbf{5.92} & \textbf{4.27} \\
\emph{Max. Accuracy Drop} & 23.55\textbf{*} & 56.70\textbf{*} / 56.36\textbf{*} & 29.90\textbf{*} & 13.62\textbf{*} &  & 22.86\textbf{*} & 56.40\textbf{*} / 57.26\textbf{*} & 32.73\textbf{*} & 13.40\textbf{*} &  & 22.40\textbf{*} & 42.76\textbf{*} / 42.91\textbf{*} & 23.23\textbf{**} & 12.38\textbf{*} \\
\emph{Semantic Similarity} & 0.84 & 0.79 / 0.77 & 0.81 & 0.86 &  & 0.85 & 0.76 / 0.75 & 0.75 & 0.87 &  & 0.82 & 0.77 / 0.78 & 0.80 & 0.85  \\
\hline

\end{tabular}
\caption{Word adversarial attack results by perturbing at most $M=3$ words. This table shows accuracy drops for various methods as well as the average semantic score of the adversarial sentences to the original sentences. \emph{Pre-attack Accuracy} represents the accuracy on original sentences. \emph{Avg. Accuracy Drop} represent the drop in accuracy averaged over the adversarial candidates, whereas \emph{Max. Accuracy Drop} represents the maximum drop in accuracy achieved if any adversarial candidate in the set of $K=20$ succeeds.
Higher the drops, more successful the attack. Here, \emph{m} and \emph{mm} represent the \emph{matched} and \emph{mismatched} versions of the dev set respectively. For Average Drops, the best results are marked in \textbf{bold} whereas the second best \underline{underlined}. For Maximum Drops, the best results are marked by \textbf{*}, while the second best by \textbf{**}.} 
\label{table:word_results}
\end{table*}

\section{Experiments}

\subsection{Datasets and Attacked Models}
We study the effectiveness of our proposed approach on various standard datasets from the well established GLUE\footnote{\footnotesize \url{https://gluebenchmark.com/}}~\citep{wang2018glue} benchmark.
It consists of the \emph{Stanford sentiment treebank (SST-2)}~\citep{socher-recursive}
(for sentiment analysis), the \emph{multi-genre natural language inference (MLNI)} corpus~\citep{williams-etal-2018-broad} and the \emph{question answering natural language inference (QNLI)} datasets~\citep{wang2018glue} (for natural language inference) and finally 
the \emph{Quora Question Pairs (QQP)} and 
the \emph{semantic textual similarity benchmark (STS-B)}~\citep{cer-semeval} (for paraphrasing).
From each dataset we pick $1000$ random instances from the dev set (except for \emph{SST-2}, where dev set size is $870$), since the actual test predictions are not publicly available.
We report the results for the classification experiments on \emph{SST-2}, \emph{MNLI}, \emph{QNLI}, and \emph{QQP} here, while for \emph{STS-B}, the results are provided in appendix.\\
We choose three dominant \emph{SOTA} models to show the efficacy of our novel \textit{target-agnostic} adversarial attack, namely \emph{BERT} ~\citep{DBLP:journals/corr/abs-1810-04805}, \emph{XLNet} ~\citep{DBLP:journals/corr/abs-1906-08237}, and \emph{BiLSTM with attention mechanism} ~\citep{wang2018glue}. A brief description of each dataset, the attacked models and the results on \emph{STS-B} are provided in the appendix.

\subsection{Attack Results}
In this section, we report the performance of the attacked models on the crafted adversarial samples from various methods. We compare our method with two well known black-box adversarial attacks methods.
The first baseline: \emph{TextFooler}~\citep{DBLP:journals/corr/abs-1907-11932}, which performs 
iterative greedy attacks heuristically by replacing the original word with a word that drops the target 
model's accuracy by the maximum margin. The other baseline is the popular \emph{genetic attack 
algorithm}~\citep{alzantot-generating}, which iteratively crafts the adversarial samples using a genetic based algorithm that fuses two adversarial sequences to create a better adversarial sample. Both these methods are run under the similar budget, word perturbation and hyperparameter settings as  our model for fair evaluation.\\
The results for word perturbation factor \emph{M=3} are provided in table \ref{table:word_results}. Hereon, we use the terms \emph{candidate set size} as well as \emph{attack budget} interchangeably, both referring to \emph{K}. \emph{Pre Attack Acc} refers to the accuracy of the models on the original sentences. \emph{Avg. Accuracy Drop} represent the drops in accuracy averaged over all the adversarial candidates, whereas \emph{Max. Accuracy  Drops} represent the maximum drop in accuracy achieved if any adversarial candidate in the set of $K=20$ succeeds. \emph{Accuracy Drops} are calculated as the difference between \emph{pre-attack} accuracy (calculated over original sentences) and \emph{post-attack} accuracy (calculated over the adversarial candidates). Hence, \emph{higher the accuracy drops, more successful the attack}. It is evident from the table that our method outperforms the baseline methods by factors of up to \textbf{15} under very limited budget settings conforming to practical attack scenarios. The semantic similarity scores of our crafted adversarial samples are generally better than those of the baselines.

We also provide a mechanism to utilize our method for \emph{character level attacks}, which makes our method all the more practical and generalized compared to methods that either employ word-specific or character-specific attacks. Results of word perturbations of $4$ and $5$, the \emph{regression} dataset \emph{STS-B} and character level attacks are provided in the appendix.

\section{Analysis}
\label{analysis_sec}
\begin{figure}[ht]
		\centering
		\includegraphics[width=60mm, height=40mm, center]{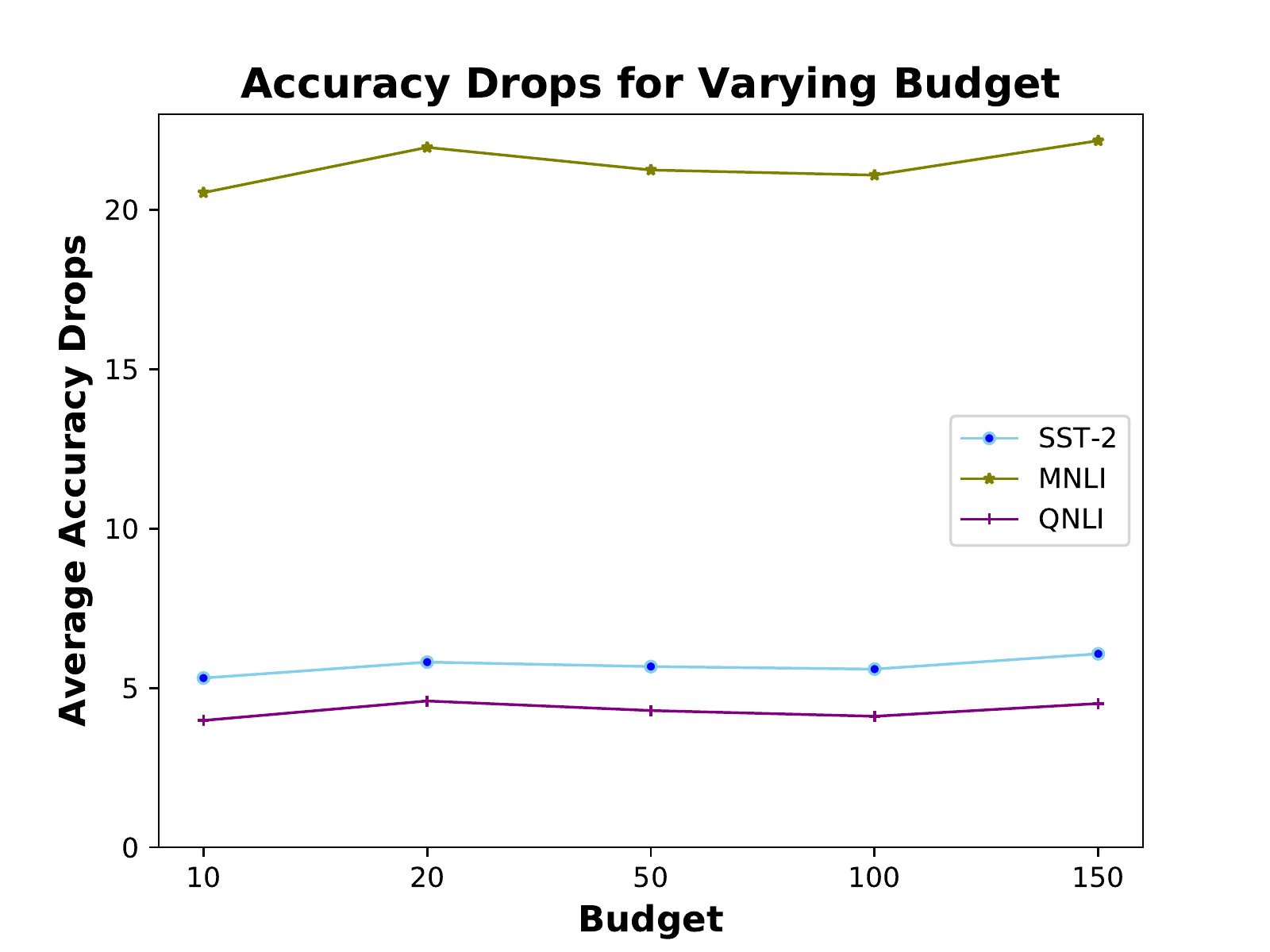}
		\caption{Average accuracy drops versus the budget value for XLNet model on various datasets.}
		\label{fig:budget_diff}
\end{figure}

\noindent\textbf{Setting the value of K:} 
We vary $K$ to ascertain the tightest possible budget for our method. 
Figure \ref{fig:budget_diff} shows the \emph{average accuracy drops} versus different \emph{budget values} for XLNet model on MNLI, QNLI, and SST-2 datasets. 
We choose the \emph{first point of inflection} at $K=20$ on all curves as our optimal value of budget $K$ across all datasets.

\noindent\textbf{Attack transferability:}
We conduct transferability experiments w.r.t to baseline methods to examine how well the adversarial samples crafted by the SOTA baselines transfer from one target model to the other. The results for \emph{QQP} and \emph{QNLI} datasets are summarized in Table~\ref{table:Transferability}, with $M$ set to $3$. In the table, \emph{higher} the value, \emph{better} the attack. For our method, we select the \emph{true}  adversarial samples per original sentence from the first model to attack the other two models. The average \% \emph{drop} in model accuracies shown in the table are  averaged over all candidates. For the baselines which follow a greedy approach, the accuracy drops are extremely low when we transfer their adversarial sentences to other models, whereas our method is much more superior in this aspect. Our method is better by factors of upto \textbf{40} (depending upon model) as compared to the baselines and can craft highly generic adversaries, while preserving (to a large extent) the \emph{semantic} and \emph{syntactic} aspects of the original sentence. Additional results for maximum drops are provided in the appendix.

\begin{table}[tbp]
		\small	
		\centering
		
		\begin{tabular}{lrrr}
			\\
			\hline
			\textbf{QQP}   &  \textbf{BERT}  & \textbf{XLNeT} &  \textbf{BiLSTM} \\
			\hline
			\emph{TextFooler Bert} & - & \underline{10.7} & \underline{12.05} \\
			\emph{TextFooler XLNeT} & 12.62\textbf{**} & - & 8.81 \\
			\hline
			\emph{Genetic Bert} & - & 0.5 & 9.19 \\
			\emph{Genetic XLNeT} & 10.68 & - & 13.58\textbf{**} \\
            \hline
			\emph{Our Bert} & - & \textbf{21.35} & \textbf{19.74} \\
			\emph{Our XLNeT} & 17.86\textbf{*} & - & 20.16\textbf{*} \\
			\hline
		\end{tabular}
		
		\begin{tabular}{lrrr}
			\\
			\hline
			\textbf{QNLI}   &  \textbf{BERT}  & \textbf{XLNeT} &  \textbf{BiLSTM} \\
			\hline
			\emph{TextFooler Bert} & - & \underline{4.33} & \underline{3.68} \\
			\emph{TextFooler XLNeT} & 0.27 & - & 1.54\textbf{**} \\
			\hline
			\emph{Genetic Bert} & - & 2.33 & 1.61 \\
			\emph{Genetic XLNeT} & 1.04\textbf{**} & - & 0.64 \\
			\hline
			\emph{Our Bert} & - & \textbf{10.28} & \textbf{9.08} \\
			\emph{Our XLNeT} & 8.68\textbf{*} & - & 9.61\textbf{*} \\
			\hline
		\end{tabular}
		
		\caption{Transferability attacks. Leftmost column header represents the original model (for which the adversaries were crafted) with the attack methods and the following three column headers represent the model on which these sentences were \emph{transferred} and evaluated. The values in the table show the average \% \emph{drop} in model accuracies on original samples and their crafted adversarial counterparts. '-' signifies same source and target model. For BERT, best results in \textbf{bold}, second best \underline{underlined}. For XLNet, best results are marked by \textbf{*}, second best by \textbf{**}.}
		\label{table:Transferability}
\end{table}

\begin{table}[ht]
		\small
		
		\centering
		\setlength\tabcolsep{3.5pt}
		\begin{tabular}{lccccc}
			\\
			\hline
			{} & \textbf{SST-2} & \textbf{MNLI(m / mm)} & \textbf{QQP} & \textbf{QNLI} \\
			\hline
			Av $\Delta$ Our & \textbf{5.81} & \textbf{21.96} / \textbf{21.48} & \textbf{6.20} & \textbf{4.59} \\
			Av $\Delta$ Random & 2.15 & 12.49 / 13.47 & 3.07 & 2.81 \\
			&\\
			\hline
			Max $\Delta$ Our & \textbf{22.86} & \textbf{56.40} / \textbf{57.26} & \textbf{32.73} & \textbf{13.40} \\			
			Max $\Delta$ Random & 14.19 & 43.86 / 42.64 & 20.46 & 8.31 \\
			
			\hline
        \end{tabular}
        \caption{Comparison of our method versus random attacks. \emph{Av $\Delta$} represents the average drop, while \emph{Max $\Delta$} represent the maximum drop  in accuracies between original and adversarial samples. Better results in bold.}
        \label{table:ablation}
\end{table}

\noindent\textbf{Ablation study:}
Since our method primarily focuses on selecting important words for crafting adversarial samples, 
we perform attacks on the target models by randomly replacing words from the 
original sentence, keeping the rest of the steps the same as provided in Section~\ref{adv_generation_section}. 
The results for the datasets with $M=3$ on XLNeT target model are provided 
in Table~\ref{table:ablation}. This table contains the average \% drop as well the maximum \% drops in accuracies between the original sentences and their adversarial counterparts caused by our method and a random attack. Once again, \textit{higher} the value achieved, \textit{better} the attack. Clearly, the proposed word importance selection criteria helps in selecting important words from the sentence and generate better adversarial samples.
\begin{table}[ht]
\scriptsize
\label{HumanEval}
\centering
\begin{tabular}{lccc}
\\
\hline
\textbf{Dataset} & \textbf{Input Type}  &  \textbf{Grammar Score}  & \textbf{Clf Consistency} \\
\hline
\multirow{2}{0.7cm}{\centering SST-2} & Original & 4.6 & \multirow{2}{0.7cm}{\centering 87.34}  \\
{} & Adversary & 4.13 &  \\
\hline
\multirow{2}{0.7cm}{\centering MNLI} & Original & 4.54 & \multirow{2}{0.7cm}{\centering 80.11} \\
{} & Adversary & 4.07 & \\
\hline
\multirow{2}{0.7cm}{\centering QQP} & Original & 4.3 & \multirow{2}{0.7cm}{\centering 86.08} \\
{} & Adversary & 4.02 & \\
            \hline
\multirow{2}{0.7cm}{\centering QNLI} & Original & 4.61 & \multirow{2}{0.7cm}{\centering 82.47} \\
{} & Adversary & 4.08 & \\
            \hline
\end{tabular}
\caption{Overall Grammatical score and Human Classification consistency for the datasets.}.
\label{table:HumanEval}
\end{table}

\noindent\textbf{Human evaluation:}
Apart from the automatic evaluation, we also analyze the adversarial samples by asking language proficient human judges to score them against the original sentences. We ask three human judges to rate the sentences based on: overall sentence structure (semantic and syntactic well-formedness) as well as consistency of the adversarial sentences against the actual sentence labels. We consider $200$ samples from each dataset for the scenario with $M=3$ perturbed words. For each original sentence, we randomly select one of the generated candidates, and randomly shuffle them with the $200$ original sentences to obtain $400$ sentences to be analyzed. The overall grammatical structure is measured on a scale of $1$--$5$, whereas for the label consistency we check the percentage of adversarial sentences that were assigned the same label as the original sentence label by human judges. The analysis is summarized in Table~\ref{table:HumanEval}. 

\begin{figure*}[tbp]
		\centering
        \begin{subfigure}{.5\textwidth}
			\includegraphics[width=80mm, height=65mm]{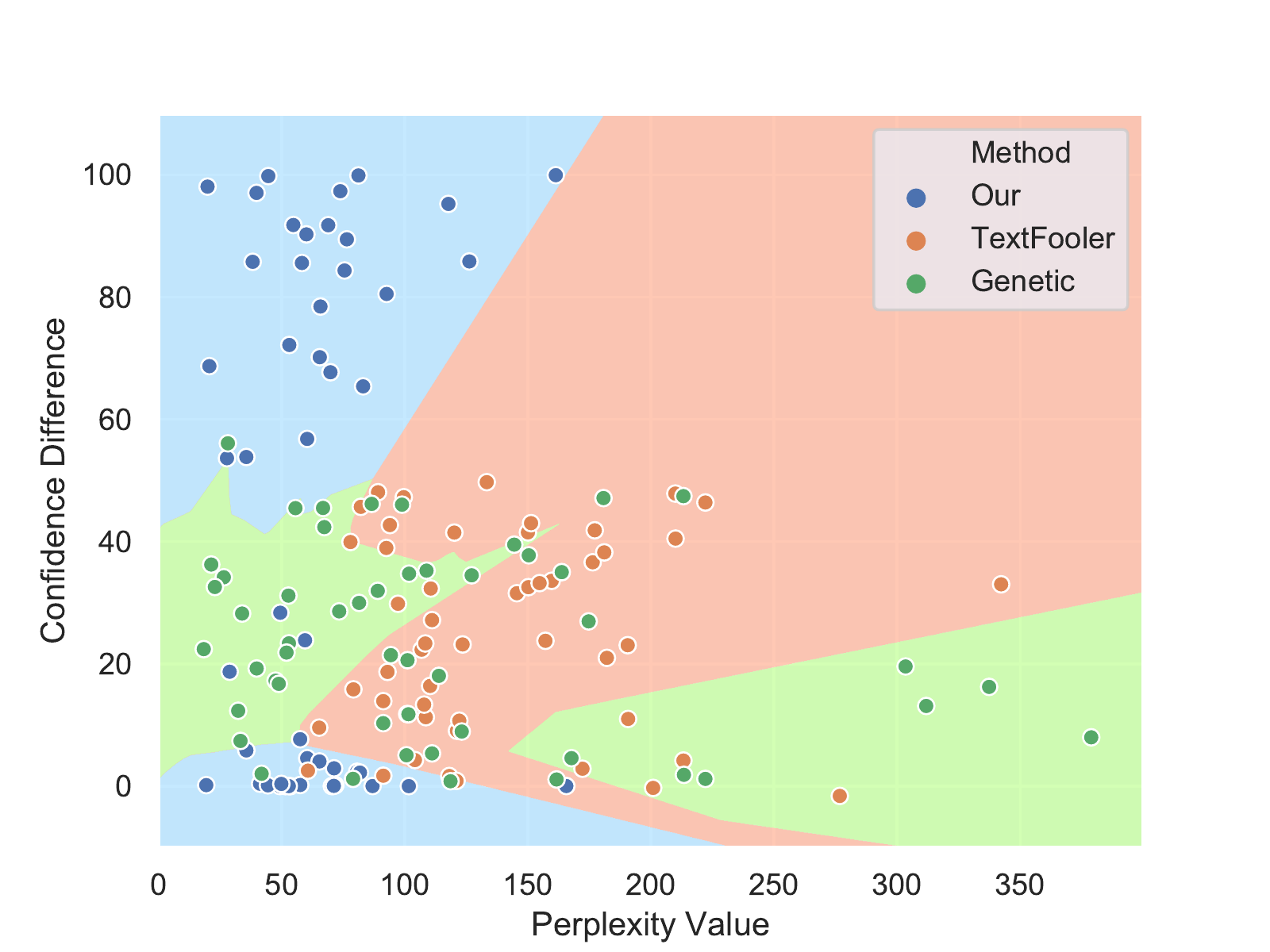}
			\caption{SST-2}
			\label{fig:sub1_bert_ppl}
		\end{subfigure}%
		\begin{subfigure}{.5\textwidth}
		
			\includegraphics[width=80mm, height=65mm]{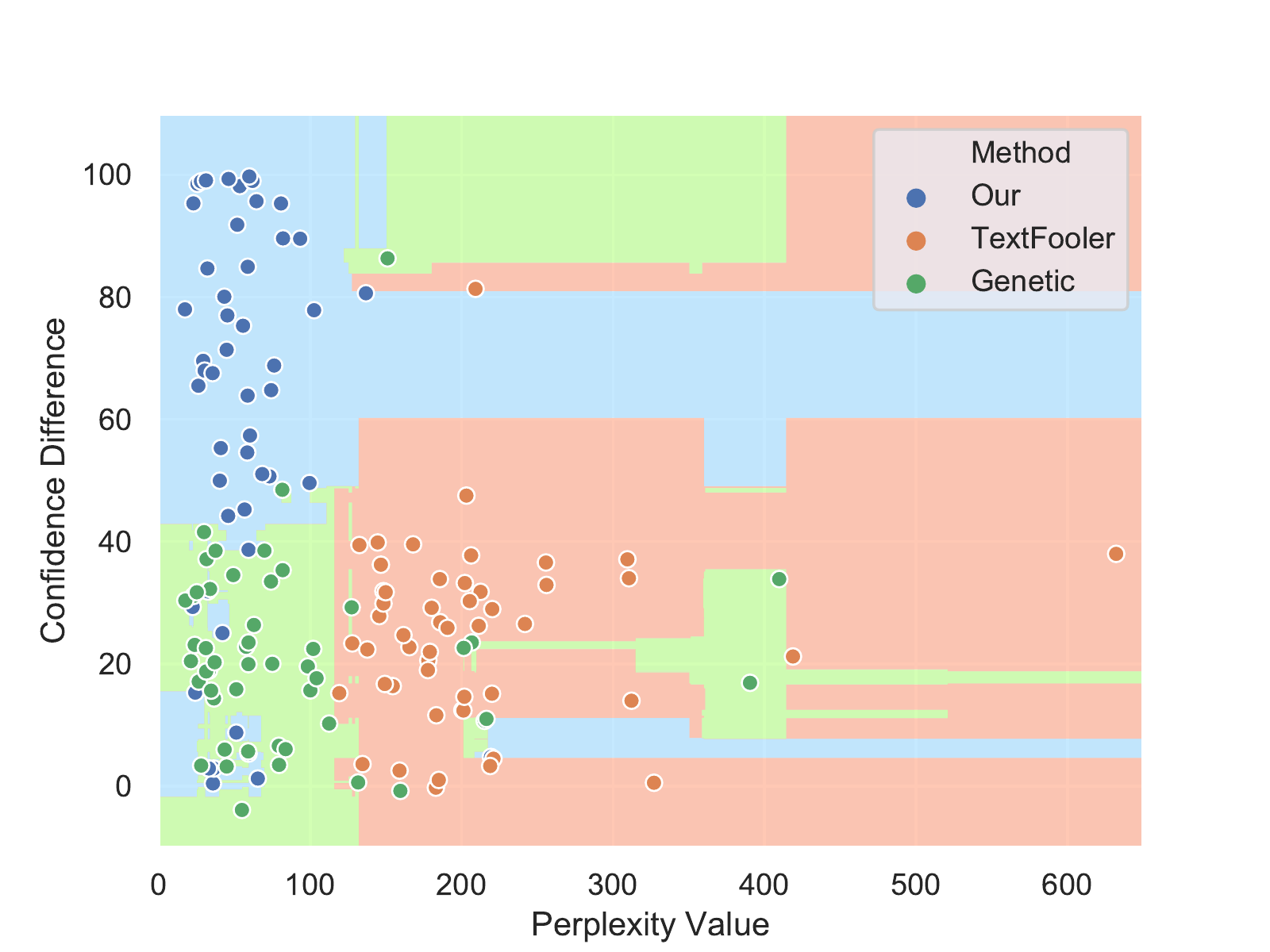}
			\caption{MNLI}
			\label{fig:sub2_xlnet_ppl}
		\end{subfigure}%
		
		\caption{Perplexity values versus confidence differences (probability difference of the correct class between the original and adversarial sentences). }
		\label{fig:ppl_scores}
\end{figure*}

\noindent\textbf{Evaluating \emph{perplexity score} versus \emph{confidence score drops} trade-off:} 
Clearly, an adversarial sentence with a very high-perplexity score has a better chance of fooling the model under attack because it is completely different from the original sentence. Ideally, we want adversarial sentences that have a low-perplexity score but can achieve high drops in accuracy.
To assess the quality of the generated adversaries and their capabilities of fooling the target models, we plot the \emph{sentence perplexity values} versus the \emph{drop in the correct class probability between the original and adversarial sentences}. Figure \ref{fig:ppl_scores} shows the plots on various datasets for $50$ randomly chosen sentences on \emph{BERT}. Observe that the majority of our adversarial sentences lie in the region of ``low-perplexity, high-confidence drops" (blue region), as opposed to baseline attacks whose adversarial sentences are spread over a very wide range of perplexity scores and are mostly confined to regions of low-to-medium confidence drops.

\begin{figure}[ht]
		\centering
			\includegraphics[width=65mm, height=50mm]{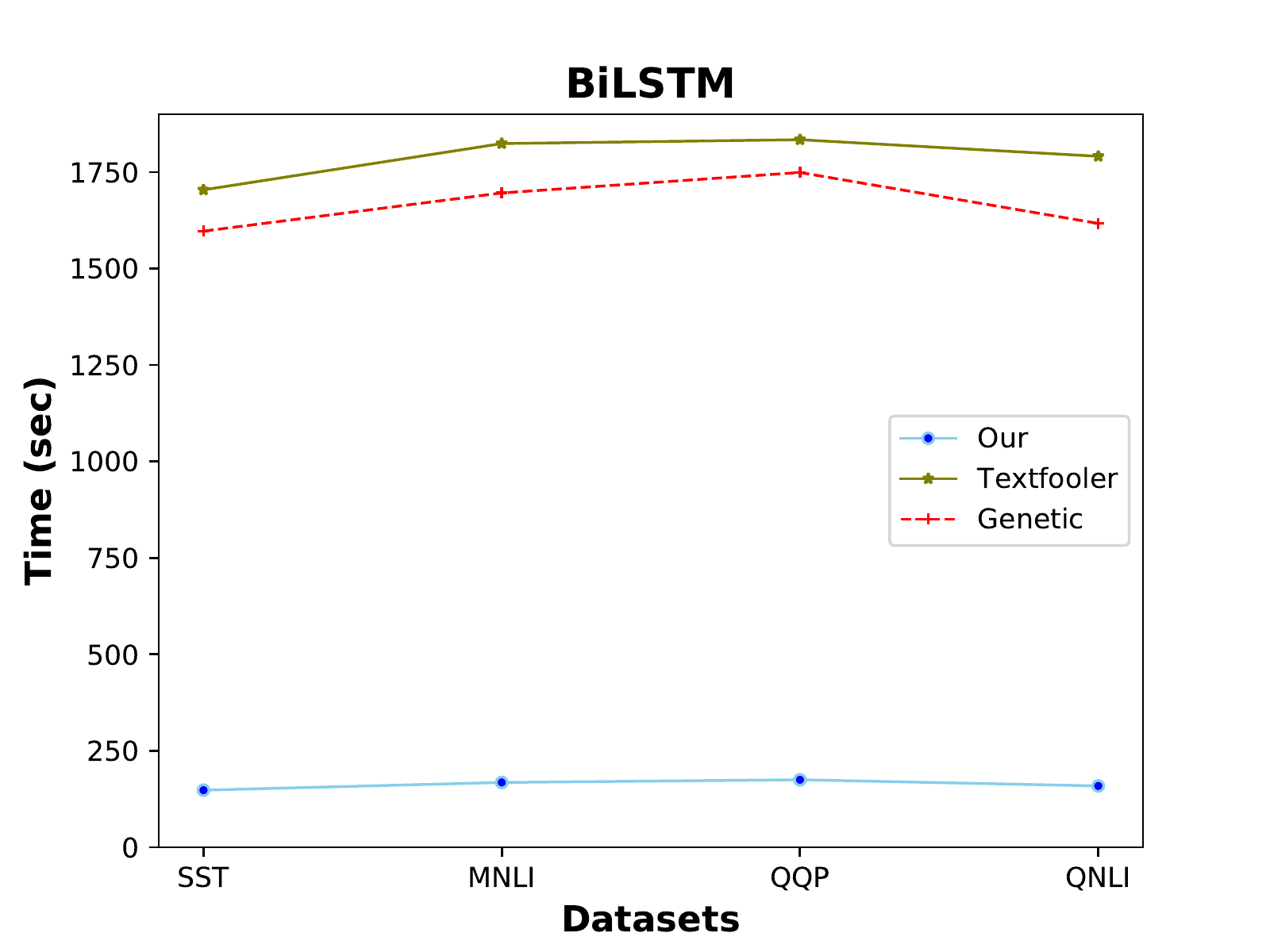}
		\caption{Runtime for the attack methods on BiLSTM target model.}
		\label{fig:runtime}
\end{figure}

\noindent\textbf{Attack time:} We analyze the time taken by different methods to attack the target models. It is evident from figure \ref{fig:runtime} that our method is faster by factors of \textbf{9-12} (depending on the dataset). We attribute this to \emph{independence} of the generated candidate adversaries from one another, thus allowing our method to perfrom attacks using mini-batches of data as opposed to the baselines, which iteratively improve the query and thus are forced to attack in a sequential manner. Similar attack run-time plots for BERT and XLNet are provided in our appendix.

\noindent\textbf{High transferability and staggered attacks:} We observed that a substantial fraction of our true adversarial examples were “common” to all models. We can thus roll out our queries per model in a “staggered approach”, i.e., querying a model, gathering its true adversarial set S and only using examples from S to target the next model, hence "shrinking" S with every model we attack. This approach can drastically reduce the attack time if we have to attack multiple models in a sequential manner (i.e, when it's not possible to attack more than one model simultaneously). Results provided in table \ref{table:Staggered_main} verify our claims. The average \textbf{\%} values shown in the table represent the  percentage of successful adversarial samples out of the total samples the corresponding model was attacked with. Further experiments have been provided in the appendix.
\begin{table}[ht]
		\centering
		\begin{tabular}{lccc}
			\\
			\hline
			\textbf{Dataset}   &  \textbf{BERT}  & \textbf{XLNet} &  \textbf{BiLSTM} \\
			\hline
			\emph{MNLI} & 34\textbf{\%} & 39\textbf{\%} & 12\textbf{\%} \\
			\emph{SST-2} & 14\textbf{\%} & 33\textbf{\%} & 17\textbf{\%} \\
			\emph{QQP} & 25\textbf{\%} & 36\textbf{\%} & 9\textbf{\%} \\
            \emph{QNLI} & 17\textbf{\%} & 41\textbf{\%} & 19\textbf{\%} \\
			\hline
		\end{tabular}
		\caption{Staggered attacks. Values in the table represent \emph{average \%} of successful adversarial samples.}
		\label{table:Staggered_main}	
\end{table}

\noindent\textbf{Comparison to previous paraphrase based attacks:} Prior paraphrase based attack methods such as ~\citep{iyyer-adversarial, ribeiro-semantically} propose template and rule based methods to curate adversaries. Verifying the correctness of these templates and rules can be difficult and may require manual investigations which is extremely time consuming. Training models to generate templates as done in ~\citep{iyyer-adversarial} requires training large scale models with large training sets, as opposed to our case. These templates and rules also have restrictions on their applicability as they can't be arbitrarily applied to any sentence. In contrast, our method is completely automated with very high semantics preservation properties, has high transferability, can attack multiple models via staggered approach as well as it has very low attack time attributed to batching of the candidates.
We also performed a quantitative and qualitative analysis of \textbf{classwise} accuracy drops for the datasets as well as show various examples of our successful adversaries, both are provided in appendix.\\

\noindent\textbf{Difference from Surrogacy based methods:} As mentioned earlier, our method follows a weaker black-box setting, where we assume the presence of a small amount of training data for ON-LSTM model. However, this is significantly different from the surrogacy based methods, where the surrogate model is explicitly trained to replicate the attacked model which requires a substantial amount of queries to the attacked model, thus incurring a huge budget.

\section{Conclusion}
We explore a novel target model agnostic adversarial attack under very limited query budgets. Unable to exploit biases of target models towards datasets, like other greedy methods do, this method carries multiple other advantages. The actual attack process is fast and the generated pool of adversarial sentences show a high degree of transferability across different types of models which brings the added benefit of attacking multiple target models simultaneously by the same candidate set. Our method performs well in comparison to the greedy SOTA adversarial attacks with a tighter budget on the number of queries, which makes our attack much more practical.



\bibliography{emnlp2020}
\bibliographystyle{acl_natbib}

\newpage

\appendix

\section{Character-Level Attacks}
Our proposed method bears another advantage attributed to the use of character level encodings at the first layer. We can thus exploit the model \textit{sensitivity} wrt each character  to final output prediction. We define character \textit{importance} by the value of the $L_2$ norm of the gradient of the classifier / regression algorithm output w.r.t the character embeddings. Furthermore, to efficiently use the important words and make better adversarial candidates, we only consider the characters from the most important words identified by the our method. We constrain the attack method to consider a maximum of $3$ words, from each of which, we consider $2$ most \textit{important} characters. These selected characters are then replaced with random characters from the standard symbols consisting of \textit{alphabets}, \textit{digits}, \textit{punctuation}, \textit{separators}, and \textit{misc symbols} like \{!, @, \#...\} etc. Since, the words formed by replacing their characters randomly may be out-of-order without any lexical structure, typically bearing misspelling errors, we thus disregard the use of a sentence similarity function, since the overall perturbation caused throughout the sentence is negligible. The results for character-level attacks are summarized in table \ref{table:char_results}.

\begin{table}[ht]
\normalsize
\centering
\begin{tabular}{lc}
\\
\hline
\textbf{Hyperparameter} & \textbf{Value} \\
\hline
\multicolumn{2}{c}{\centering \bf Character CNN}\\
\cline{1-2}\\
Embedding dims & 100 \\
CNN kernel sizes & 3,4,5\\
CNN Channels & 100 \\
dropout & 0.3 \\
Char CNN Output Dims & 200 \\
\hline

\multicolumn{2}{c}{\centering \bf ON-LSTM}\\
\hline
Num Layers & 2\\
Chunk Size & 10 \\
Hidden dims & 500\\
dropout & 0.25 \\
\hline

\multicolumn{2}{c}{\centering \bf Attention Blocks}\\
\hline
\# Heads: MultiHead Att & 4\\
$C^{Att}$ dims & 100, 1\\
$C^{clf}$ dims & 300, 100\\
\hline

\multicolumn{2}{c}{\centering \bf Training Parameters}\\
\hline
Default lr & $1e^{-3}$ \\
Max Epochs & 200\\
Batch Size & 32 \\
Optimizer & Adam \\

\hline
\end{tabular}
\caption{ON-LSTM Training Hyperparameters}.
\label{table:lstm_hyp}
\end{table}

\section{Model Details}
Here we provide the details of hyper-parameters of the enhanced On-LSTM model and our adversarial attack method. The general hyper-parameters which were kept same for all tasks are provided in Table \ref{table:lstm_hyp}, while remaining hyper-parameters like - weight decay, learning rate etc. were adjusted specifically to tasks with appropriate scheduling of learning rate on plateau. The outputs embeddings from the character CNN are normalized before inputting into the LSTM. \\
The value of \textit{k} while choosing $k$-most-semantically-similar words for each token of the original sentence is set to $25$. The threshold $\phi$ to drop neighbors below a certain cosine score is set to $0.5$, whereas the \emph{sentence semantic similarity} threshold $\epsilon$ is set to $0.5$ as well. The parameter $W$, referring to the number of sentences chosen after passing through the On-LSTM model in increasing order of perplexity scores is set to $350$.

\section{Attack Transferability}
More comparisons for attack transferability across targeted models are provided in the  tables \ref{table:Transferability_3word_qqp}, \ref{table:Transferability_3word_qnli} and \ref{table:Transferability_4word_qnli}. The values in the tables show the \textbf{average \% drops} in model accuracies on original samples and their crafted adversarial counterparts. Thus \emph{higher} the value, \emph{better} is the attack. The leftmost column represents the models(for which the adversarial sentences were crafted) along with the attack methods and the following 3 columns represent the model on which these sentences were evaluated. '-' signifies same source and target model.

\begin{table}[ht]
		\small
		\centering
		\begin{tabular}{lccc}
			\\
			\hline
			\textbf{QQP}   &  \textbf{BERT}  & \textbf{XLNeT} &  \textbf{BiLSTM} \\
			\hline
			
			\emph{TextFooler Bert} & - & \underline{14.0} & \underline{17.56} \\
			\emph{TextFooler XLNeT} & 16.03\textbf{**} & - & 11.41 \\
			\hline
			\emph{Genetic Bert} & - & 1.19 & 12.52 \\
			\emph{Genetic XLNeT} & 15.13 & - & 20.19\textbf{**} \\
            \hline
            \emph{Our Bert} & - & \textbf{28.34} & \textbf{25.48} \\
			\emph{Our XLNeT} & 24.93\textbf{*} & - & 26.07\textbf{*} \\
			\hline
		\end{tabular}
		
		\centering
		\begin{tabular}{lccc}
			\\
			\hline
			\textbf{QNLI}   &  \textbf{BERT}  & \textbf{XLNeT} &  \textbf{BiLSTM} \\
			\hline
			
			\emph{TextFooler Bert} & - & \underline{5.89} & \underline{5.3} \\
			\emph{TextFooler XLNeT} & 0.56 & - & 2.9\textbf{**} \\
			\hline
			\emph{Genetic Bert} & - & 4.68 & 3.33 \\
			\emph{Genetic XLNeT} & 1.65\textbf{**} & - & 1.14 \\
			\hline
			\emph{Our Bert} & - & \textbf{15.31} & \textbf{14.35} \\
			\emph{Our XLNeT} & 13.16\textbf{*} & - & 12.98\textbf{*} \\
			\hline
		\end{tabular}
		
		\caption{Attack transferability on QQP and QNLI with maximum allowed perturbation of upto 4 words. For BERT, best results in \textbf{bold}, second best \underline{underlined}. For XLNet, best results are marked by \textbf{*}, second best by \textbf{**}.}
		\label{table:Transferability_3word_qqp}

\end{table}

\begin{table}[ht]
		\small
		
		\centering
		\begin{tabular}{lccc}
			\\
			\hline
			\textbf{SST-2}   &  \textbf{BERT}  & \textbf{XLNeT} &  \textbf{BiLSTM} \\
			\hline
			
			\emph{TextFooler Bert} & - & 17.78 & \underline{16.33} \\
			\emph{TextFooler XLNeT} & 19.71\textbf{**} & - & 17.19 \\
			\hline
			\emph{Genetic Bert} & - & \underline{20.01} & 14.99 \\
			\emph{Genetic XLNeT} & 18.74 & - & 17.80\textbf{**} \\
			\hline
			\emph{Our Bert} & - & \textbf{26.67} & \textbf{28.92} \\
			\emph{Our XLNeT} & 25.79\textbf{*} & - & 27.06\textbf{*} \\
			\hline
		\end{tabular}
		
		\setlength\tabcolsep{2.5pt}
		\begin{tabular}{lccc}
			\\
			\hline
			\textbf{MNLI}   &  \textbf{BERT}  & \textbf{XLNeT} &  \textbf{BiLSTM} \\
			\hline
			
			\emph{TF Bert} & - & \underline{18.14} / \underline{19.06} & \underline{16.31} / \underline{15.21} \\
			\emph{TF XLNeT} & 21.86\textbf{/}20.51\textbf{**} & - & 17.16\textbf{/}16.63\textbf{**} \\
			\hline
			\emph{Genetic Bert} & - & 3.54 / 3.87 & 3.47 / 3.14 \\
			\emph{Genetic XLNeT} & 6.04 / 6.06 & - & 6.02 / 4.87 \\
            \hline
            \emph{Our Bert} & - & \textbf{36.81} / \textbf{36.08} & \textbf{35.25} / \textbf{37.09} \\
			\emph{Our XLNeT} & 38.07 / 37.35\textbf{*} & - & 39.05 / 40.22\textbf{*} \\
			\hline
		\end{tabular}
		
		\caption{Attack transferability on SST-2 and MNLI(Matched/MisMatched) with maximum allowed perturbation of upto 3 words. TF stands for the baseline \textit{TextFooler}. For BERT, best results in \textbf{bold}, second best \underline{underlined}. For XLNet, best results are marked by \textbf{*}, second best by \textbf{**}.}
		\label{table:Transferability_3word_qnli}
		
\end{table}

\begin{table}[ht]
		\small
		
		\centering
		
		\begin{tabular}{lccc}
			\\
			\hline
			\textbf{SST-2}   &  \textbf{BERT}  & \textbf{XLNeT} &  \textbf{BiLSTM} \\
			\hline
			
			\emph{TextFooler Bert} & - & 25.75 & 23.29 \\
			\emph{TextFooler XLNeT} & 18.59 & - & 22.36\textbf{**} \\
			\hline
			\emph{Genetic Bert} & - & \underline{26.44} & \underline{30.86} \\
			\emph{Genetic XLNeT} & 23.33\textbf{**} & - & 21.72 \\
			\hline
			\emph{Our Bert} & - & \textbf{39.22} & \textbf{37.39} \\
			\emph{Our XLNeT} & 32.62\textbf{*} & - & 34.08\textbf{*} \\
			\hline
		\end{tabular}
		
		\setlength\tabcolsep{2.5pt}
		\begin{tabular}{lccc}
			\\
			\hline
			\textbf{MNLI}   &  \textbf{BERT}  & \textbf{XLNeT} &  \textbf{BiLSTM} \\
			\hline
			
			\emph{TF Bert} & - & \underline{25.34} / \underline{25.44} & \underline{22.47} / \underline{24.04} \\
			\emph{TF XLNeT} & 26.7 / 28.53\textbf{**} & - & 25.33 / 25.06\textbf{**} \\
			\hline
			\emph{Genetic Bert} & - & 8.14 / 9.03 & 6.49 / 7.13 \\
			\emph{Genetic XLNeT} & 12.7 / 11.73 & - & 10.64 / 10.06 \\
            \hline
            \emph{Our Bert} & - & \textbf{46.87} / \textbf{49.06} & \textbf{50.08} / \textbf{48.68} \\
			\emph{Our XLNeT} & 47.81 / 49.63\textbf{*} & - & 49.02 / 47.93\textbf{*} \\
			\hline
		\end{tabular}
		
		\caption{Attack transferability on SST-2 and MNLI(Matched/MisMatched) with maximum allowed perturbation of upto 4 words. TF stands for the baseline \textit{TextFooler}. For BERT, best results in \textbf{bold}, second best \underline{underlined}. For XLNet, best results are marked by \textbf{*}, second best by \textbf{**}.}
		\label{table:Transferability_4word_qnli}
		
\end{table}

\section{Attack Time Comparison}
\begin{figure*}[ht]
		\centering
        \begin{subfigure}{.50\textwidth}
			\includegraphics[width=80mm, height=60mm]{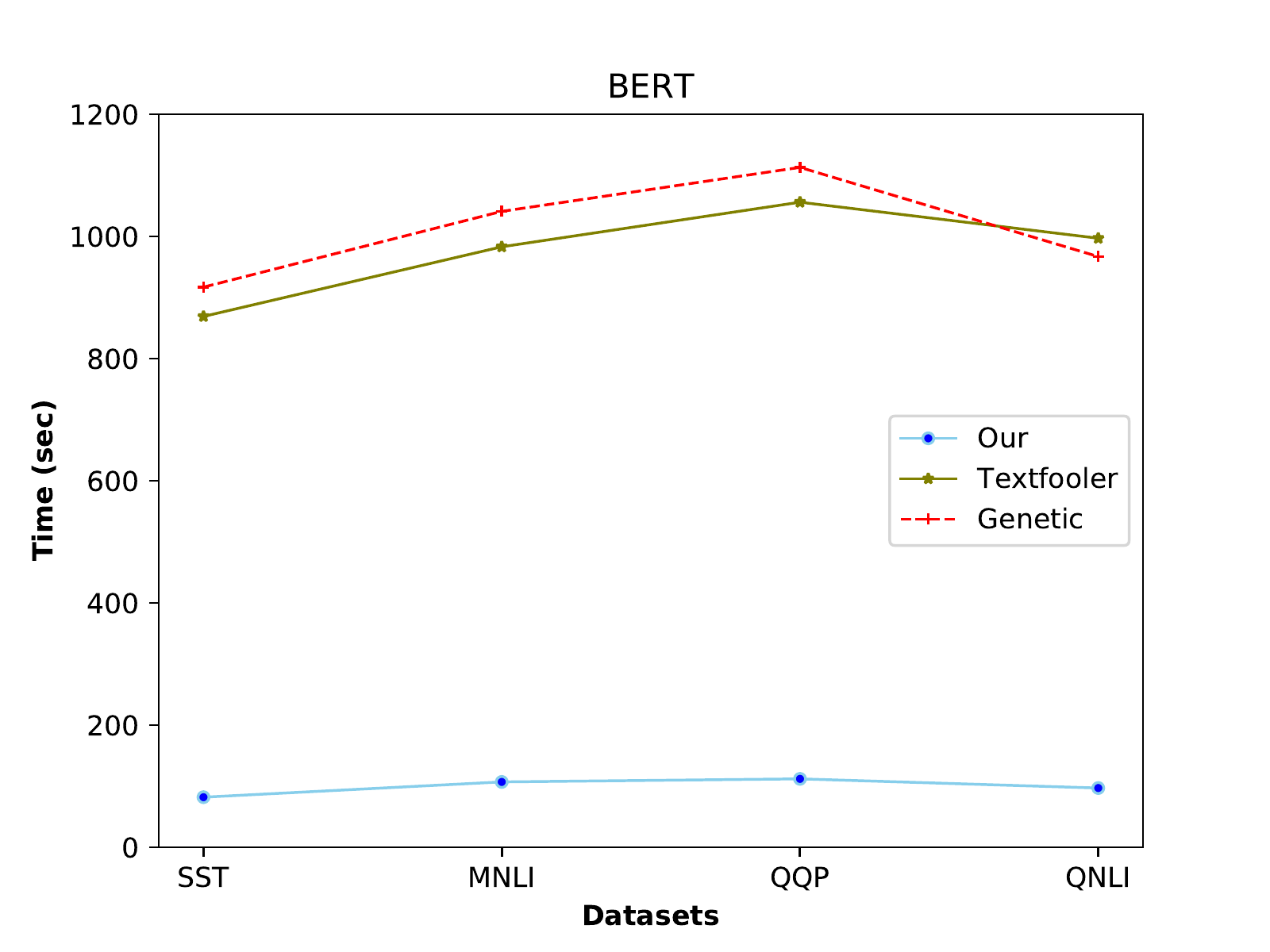}
			\caption{Attack Time on BERT}
			\label{fig:sub1_bert}
		\end{subfigure}%
		\begin{subfigure}{.50\textwidth}
			\centering
			\includegraphics[width=80mm, height=60mm]{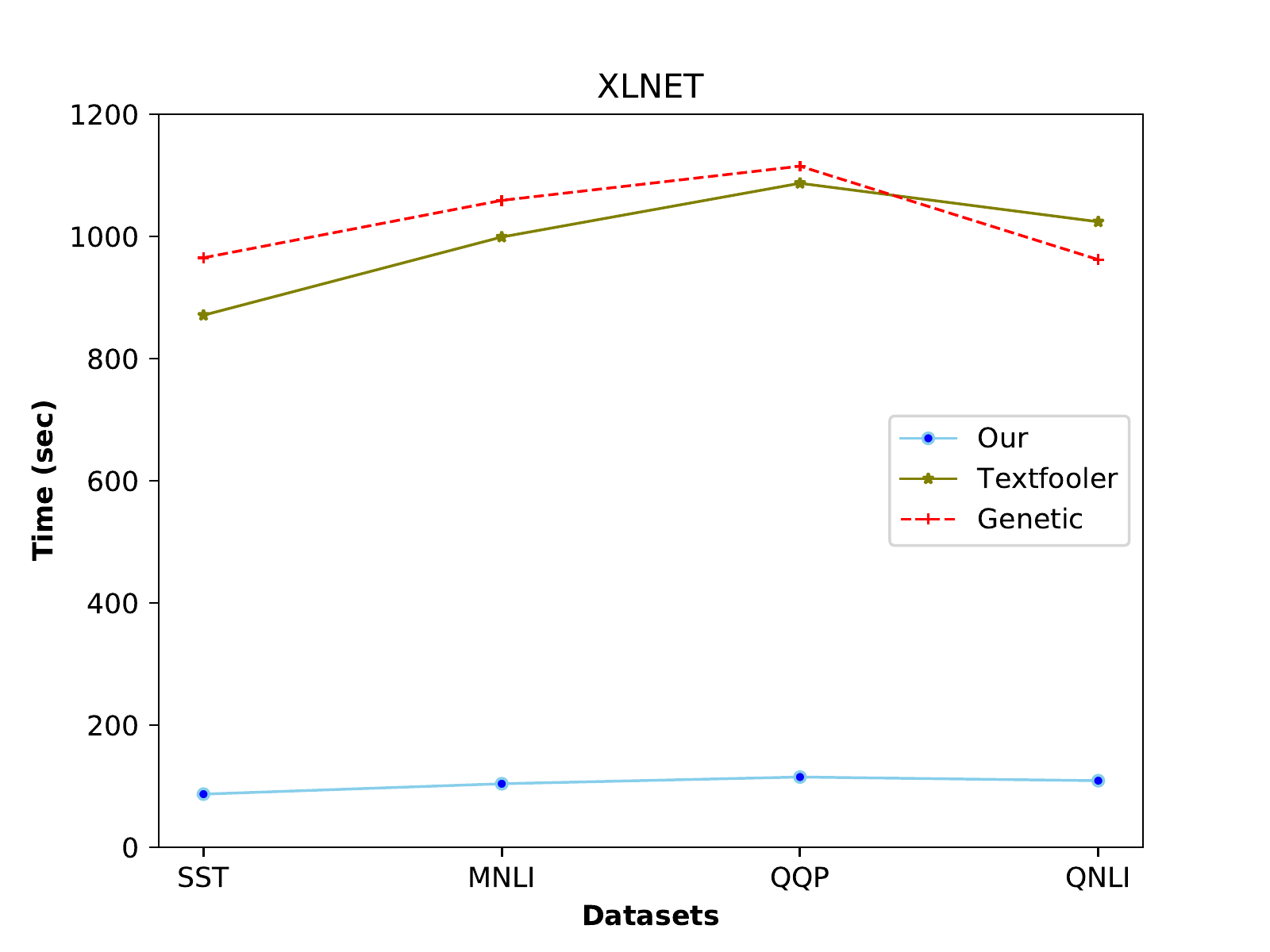}
			\caption{Attack Time on XLNET}
			\label{fig:sub2_xlnet}
		\end{subfigure}%
		
		\caption{Runtime for the attack methods.}
		\label{fig:runtime_ap}
\end{figure*}

Figure \ref{fig:runtime_ap} shows the time taken by various methods to attack the target models: \emph{BERT and XLNet}. Its evident that our method is faster by factors of \textbf{8-12} (depending upon the dataset). 

\newpage
\section{High transferability and Staggered Attacks}
Table \ref{table:Staggered} provide more results for the \emph{Staggered} attack approach. The average \textbf{\%} values shown in the table represent the  percentage of successful adversarial samples out of the total samples the corresponding model was attacked with. The rate is high when successful adversaries are transferred from \emph{BERT} to \emph{XLNet} and vice-versa which we believe is due to the model architecture bias. The first model is attacked with the budget of $K=20$, the successful adversaries from this model are then used to attack the second model and the process repeats, thus shrinking the size of the set of adversarial candidates.

\begin{table}[ht]
		\small
		
		\centering
		\begin{tabular}{lccc}
			\\
			\hline
			\textbf{Dataset}   &  \textbf{XLNet}  & \textbf{BERT} &  \textbf{BiLSTM} \\
			\hline
			\emph{MNLI} & 36\textbf{\%} & 38\textbf{\%} & 11\textbf{\%} \\
			\emph{SST-2} & 13\textbf{\%} & 29\textbf{\%} & 14\textbf{\%} \\
			\emph{QQP} & 22\textbf{\%} & 18\textbf{\%} & 6\textbf{\%} \\
            \emph{QNLI} & 21\textbf{\%} & 34\textbf{\%} & 16\textbf{\%} \\
			\hline
		\end{tabular}
		
		\centering
		\begin{tabular}{lccc}
			\\
			\hline
			\textbf{Dataset}   &  \textbf{BiLSTM}  & \textbf{XLNet} &  \textbf{BERT} \\
			\hline
			\emph{MNLI} & 42\textbf{\%} & 24\textbf{\%} & 45\textbf{\%} \\
			\emph{SST-2} & 16\textbf{\%} & 13\textbf{\%} & 30\textbf{\%} \\
			\emph{QQP} &  23\textbf{\%} & 19\textbf{\%} & 37\textbf{\%} \\
            \emph{QNLI} & 9\textbf{\%} & 10\textbf{\%} & 28\textbf{\%} \\
			\hline
		\end{tabular}
		
		\caption{Staggered attacks. Values in the table represent \emph{average \%} of successful adversarial samples.}
		\label{table:Staggered}
		
\end{table}

\begin{table*}[ht]
\tiny

\centering
\setlength\tabcolsep{3.5pt}
\begin{tabular}{lcccccccccccccc}
\\
\hline
\textbf{}   &  \multicolumn{5}{c}{\bf \emph{BERT}}  &  \multicolumn{5}{c}{\bf \emph{XLNeT}} & \multicolumn{4}{c}{\bf \emph{BiLSTM + Attn + ELMo}} \\
\cline{2-5}   \cline{7-10} \cline{12-15}
&  SST-2  & MNLI(m / mm) & QQP  & QNLI &  &  SST-2  & MNLI(m / mm) & QQP  & QNLI &  &  SST-2  & MNLI(m / mm) & QQP  & QNLI \\

\hline
\emph{Pre Attack Acc.} & 92.14 & 82.7 / 84.88 & 90.5 & 84.65 &  & 93.3 & 85.00 / 85.38 & 88.93 & 83.21 &  & 88.68 & 70.1 / 71.36 & 81.4 & 76.2 \\                                                      

&\\

\hline
\textbf{} &  \multicolumn{14}{c}{\bf \emph{2 Characters perturbed in 1 Word}} \\
\hline
\emph{Avg. Accuracy Drops} & 2.95 & 20.92 / 21.14 & 8.31 & 2.02 & & 3.05 & 21.15 / 20.93 & 9.82 & 1.83 &  & 3.13 & 19.96 / 20.80 & 9.24 & 2.37 \\
\emph{Max. Accuracy Drops} & 8.79 & 42.91 / 42.25 & 26.82 & 7.11 &  & 8.93 & 43.81 / 42.90 & 27.05 & 6.88 &  & 11.52 & 34.82 / 33.81 & 22.51 & 8.09 \\
\emph{Semantic Sim} & 0.89 & 0.86 / 0.85 & 0.88 & 0.90 &  & 0.89 & 0.84 / 0.83 & 0.82 & 0.91 &  & 0.88 & 0.83 / 0.84 & 0.87 & 0.92 \\

&\\

\hline           
\textbf{} &  \multicolumn{14}{c}{\bf \emph{2 Characters perturbed in 2 Words}} \\
\hline
\emph{Avg. Accuracy Drops} & 6.66 & 26.67 / 25.82 & 15.24 & 3.29 &  & 7.21 & 28.63  / 28.07 & 16.32 & 3.09 &  & 6.38 & 25.61 / 26.79 & 14.82 & 3.92 \\
\emph{Max. Accuracy Drops} & 15.81 & 49.82 / 48.72 & 28.16 & 11.19 &  & 15.85 & 50.34 / 51.07 & 29.04 & 10.50 &  & 14.80 & 43.56 / 44.08 & 24.67 & 12.41 \\
\emph{Semantic Sim} & 0.83 & 0.81 / 0.82 & 0.83 & 0.84 &  & 0.83 & 0.79 / 0.78 & 0.77 & 0.84 &  & 0.81 & 0.77 / 0.76 & 0.81 & 0.85  \\
&\\

\hline
\textbf{} &  \multicolumn{14}{c}{\bf \emph{2 Characters perturbed in 3 Words}} \\
\hline
\emph{Avg. Accuracy Drops} & 10.65 & 33.56 / 34.18 & 18.63 & 4.56 &  & 11.62 & 34.82 / 35.08 & 19.74 & 4.28 &  & 9.83 & 29.75 / 29.07 & 17.22 & 5.42 \\
\emph{Max. Accuracy Drops} & 21.32 & 58.95 / 57.24 & 29.91 & 13.96 &  & 22.52 & 56.87 / 57.23 & 30.82 & 12.97 &  & 18.89 & 49.74 / 48.08 & 26.83 & 14.23 \\
\emph{Semantic Sim} & 0.79 & 0.75 / 0.74 & 0.77 & 0.78 &  & 0.76 & 0.72 / 0.73 & 0.71 & 0.79 &  & 0.75 & 0.71 / 0.70 & 0.74 & 0.79 \\
\hline

\end{tabular}
\caption{Character level adversarial attack results of our method by perturbing $2$ characters in \textbf{1}, \textbf{2} and \textbf{3} words respectively. \emph{Pre Attack Acc} represents the accuracy on original sentences. \emph{Avg. Accuracy Drops} represent the drops in accuracy averaged over the adversarial candidates, whereas \emph{Max. Accuracy Drops} represent the maximum drop in accuracy achieved if any adversarial candidate in the set of $K=20$ succeeds. \emph{Semantic Sim} represents the average semantic similarity of the adversaries to the original sentences. Here, \emph{m} and \emph{mm} represent the \emph{matched} and \emph{mismatched} versions of the dev set respectively.} 
\label{table:char_results}
\end{table*}

\section{Classwise Performance} 
\begin{figure*}[t]
		\centering
        \begin{subfigure}{.35\textwidth}
			\includegraphics[width=60mm, height=40mm]{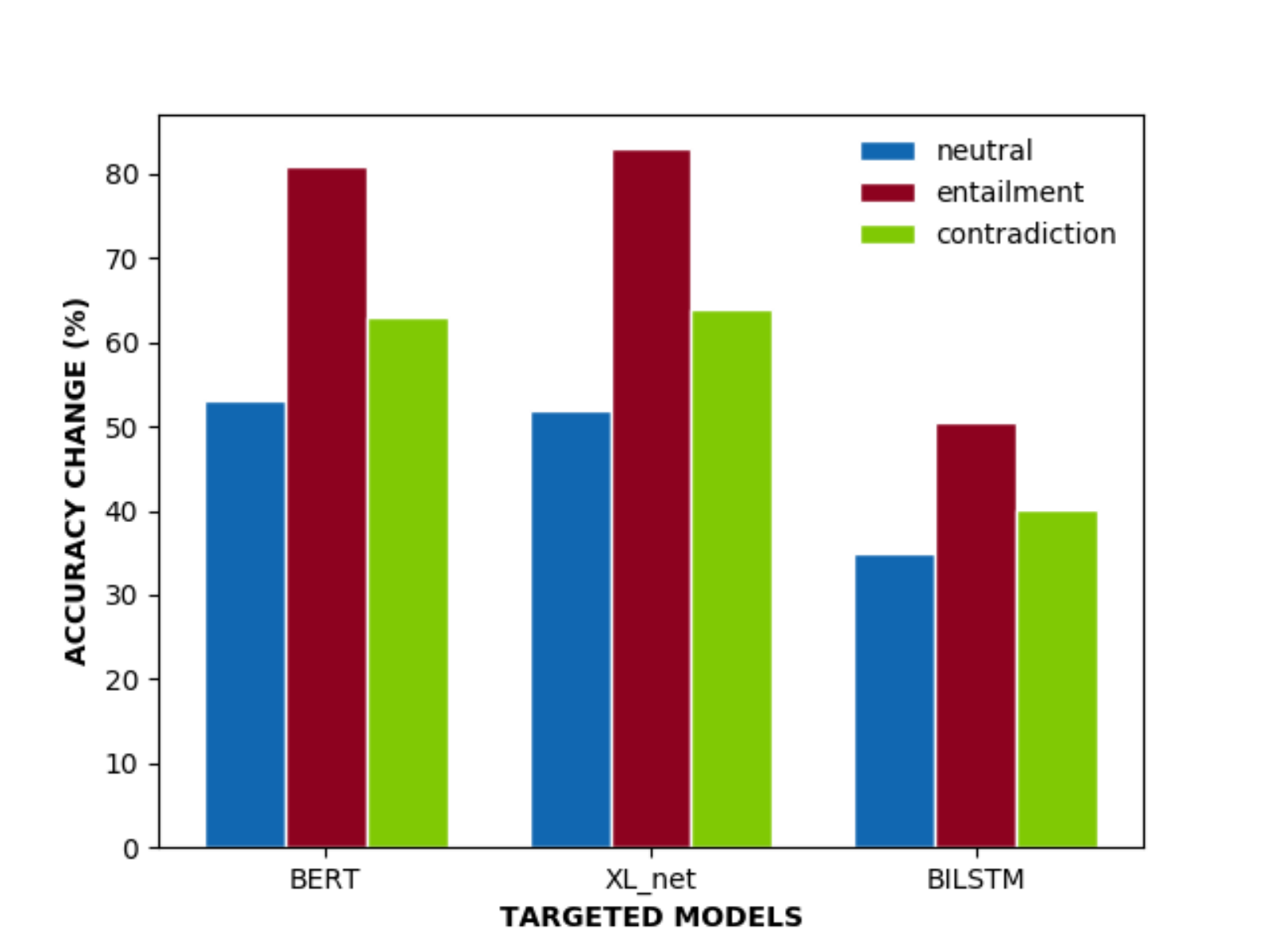}
			\caption{MNLI}
		\end{subfigure}%
		\begin{subfigure}{.32\textwidth}
			\centering
			\includegraphics[width=55mm, height=40mm]{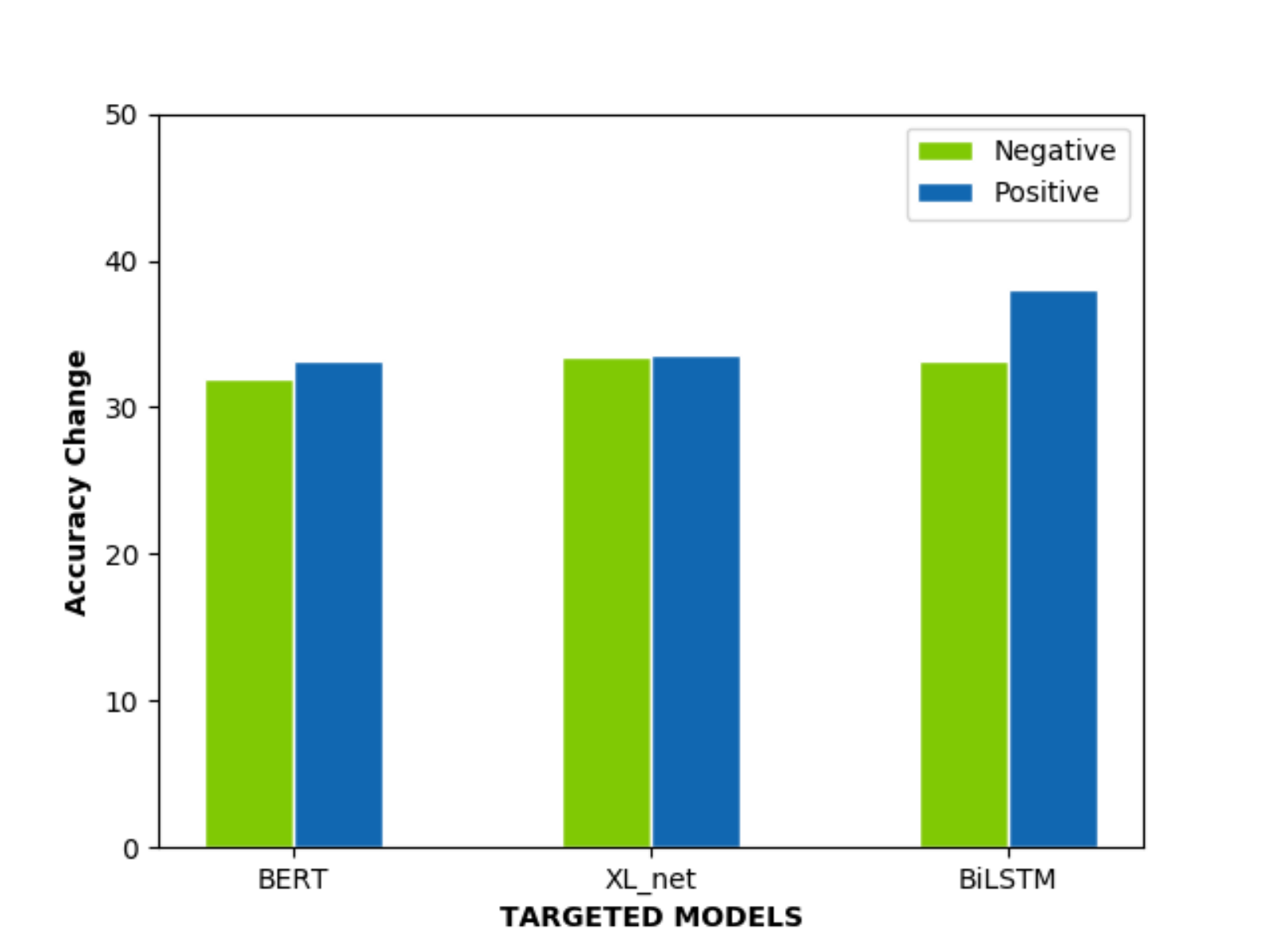}
			\caption{SST-2}
		\end{subfigure}%
		\begin{subfigure}{.32\textwidth}
			\centering
			\includegraphics[width=55mm, height=40mm]{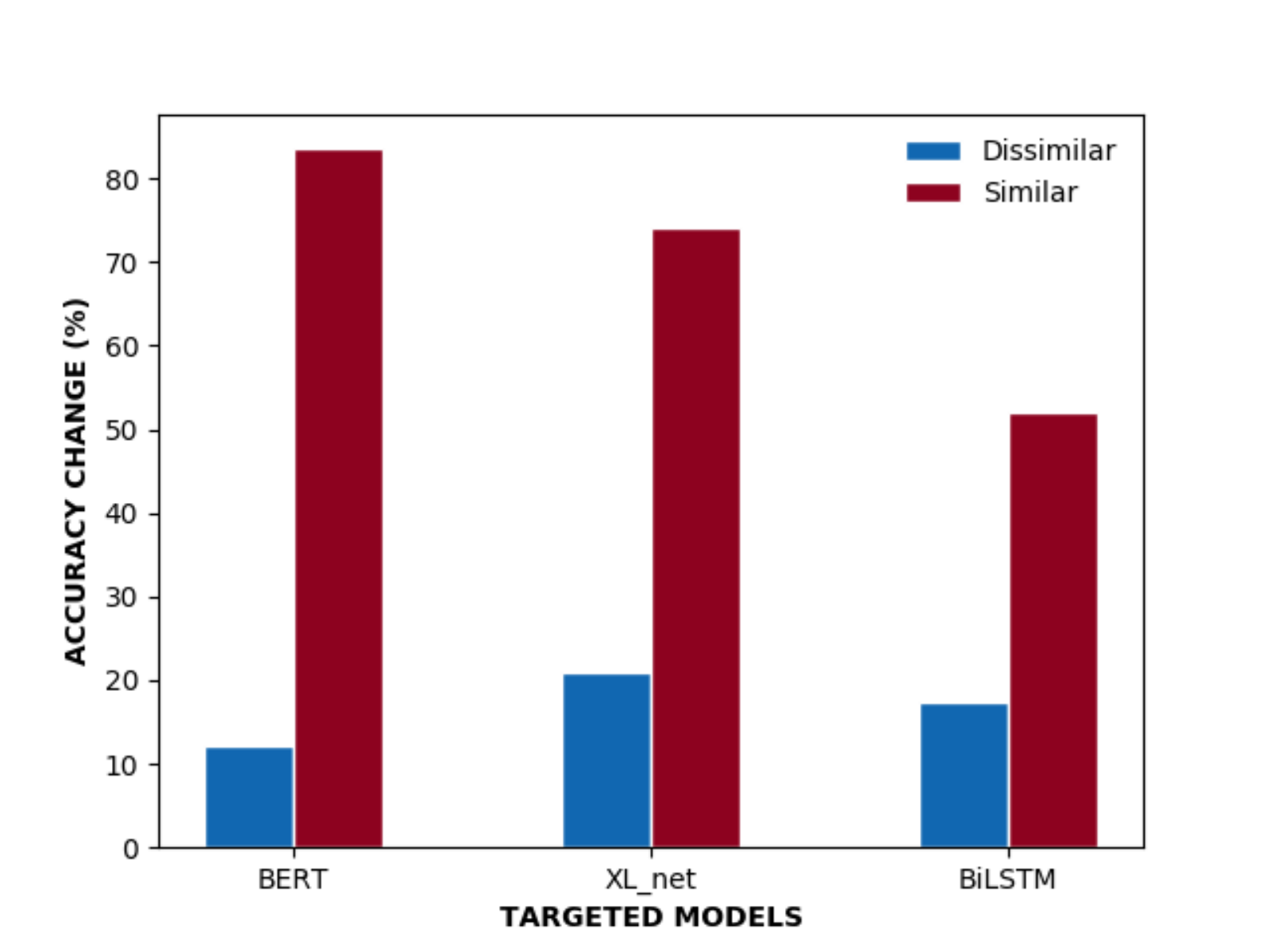}
			\caption{QQP}
		\end{subfigure}
		\caption{Classwise accuracy change for MNLI, SST-2 and QQP on the three target models with maximum perturbation allowed upto 3 words.}
		\label{fig:Class_acc_diff_3_word}
\end{figure*}

\begin{figure*}[ht]
    \centering
		\begin{subfigure}{.35\textwidth}
			\includegraphics[width=50mm, height=30mm]{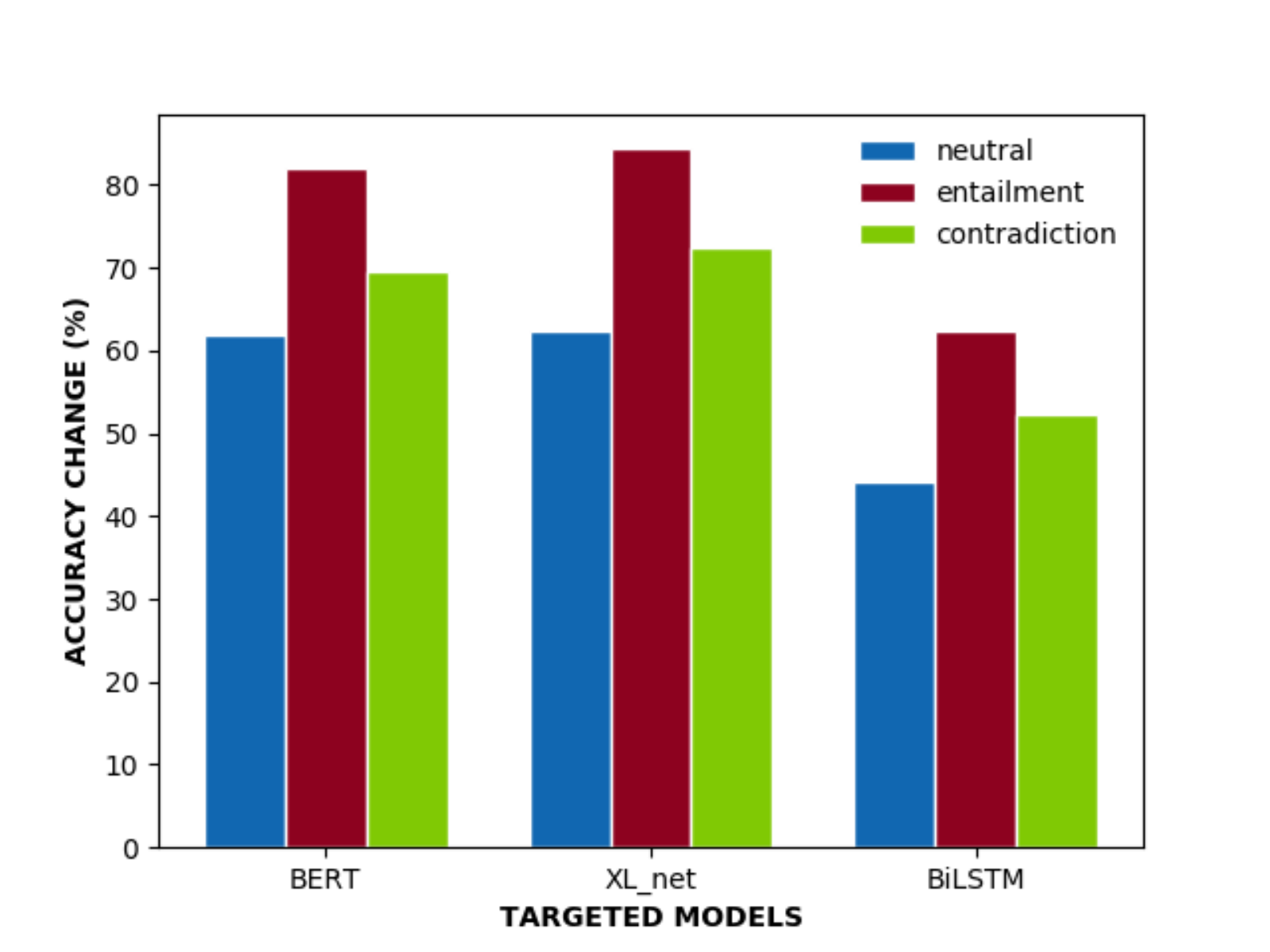}
			\caption{MNLI}
		\end{subfigure}%
		\begin{subfigure}{.32\textwidth}
			\centering
			\includegraphics[width=45mm, height=30mm]{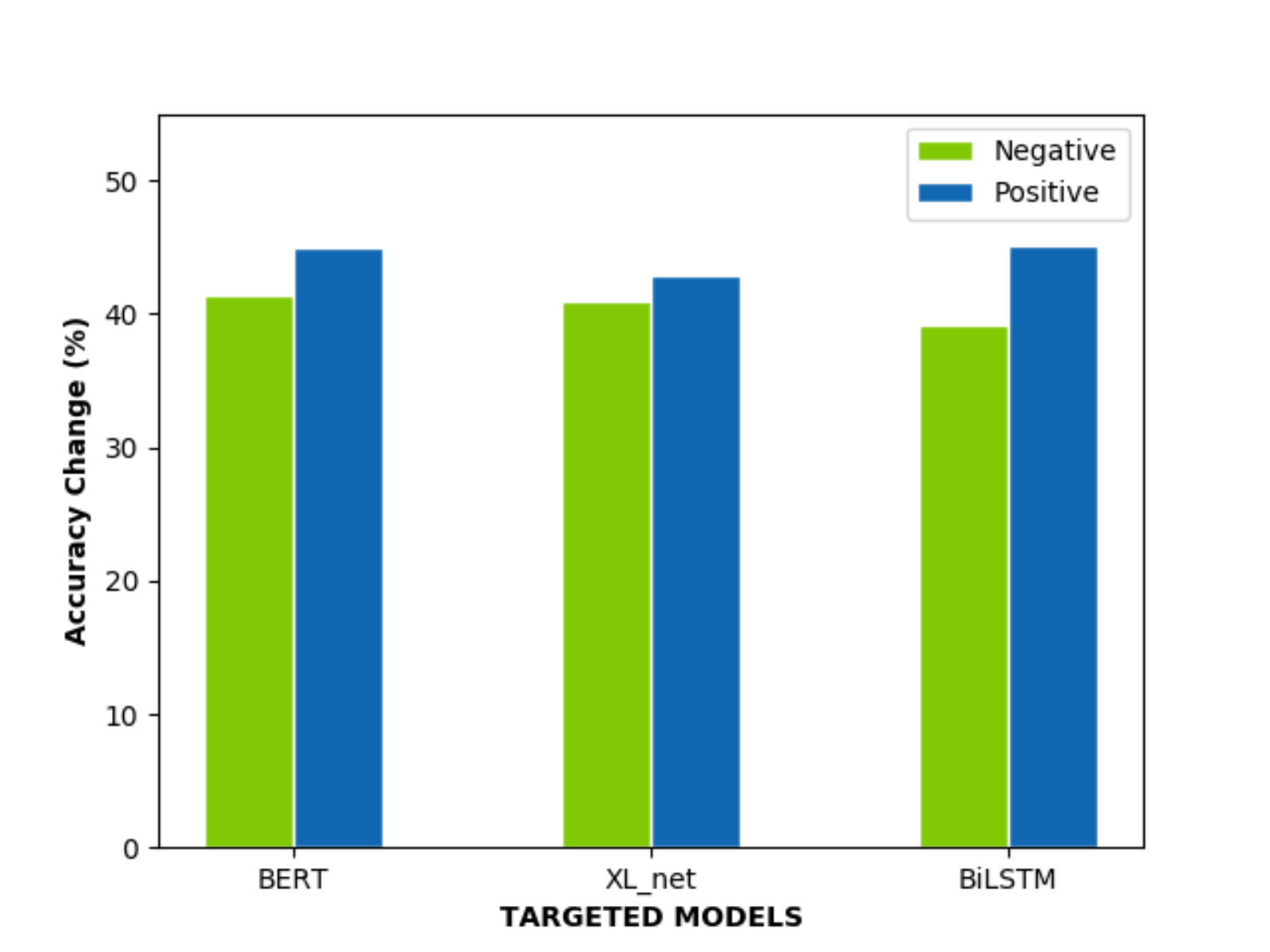}
			\caption{SST-2}
		\end{subfigure}%
		\begin{subfigure}{.32\textwidth}
			\centering
			\includegraphics[width=45mm, height=30mm]{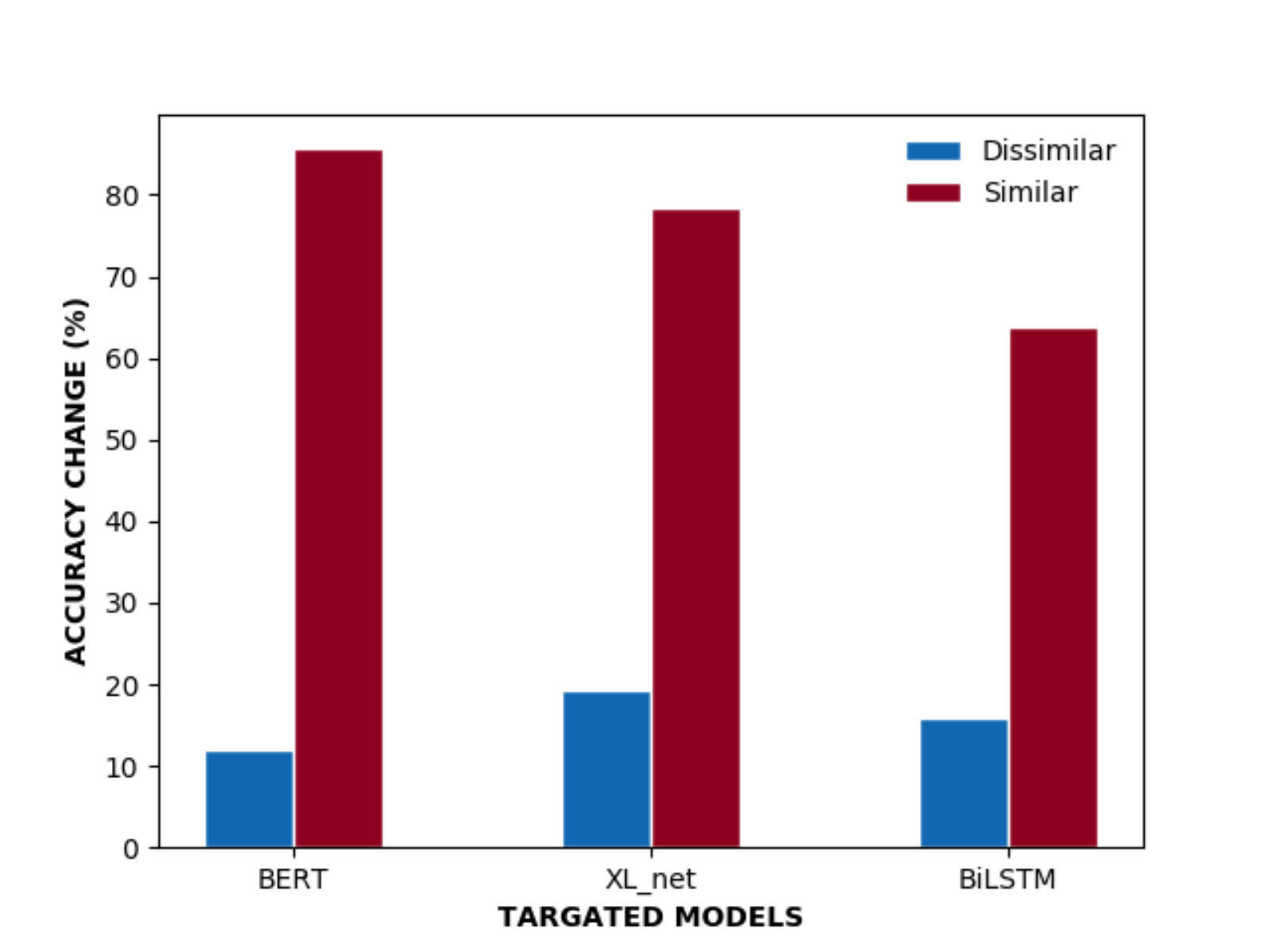}
			\caption{QQP}
		\end{subfigure}
		\caption{Classwise accuracy change for MNLI, SST-2 and QQP on the three target models with maximum perturbation allowed upto 4 words.}
		\label{fig:Class_acc_diff_4_word}
\end{figure*}

\begin{figure*}[ht]
		\centering
        \begin{subfigure}{.5\textwidth}
            \centering
			\includegraphics[width=60mm, height=40mm]{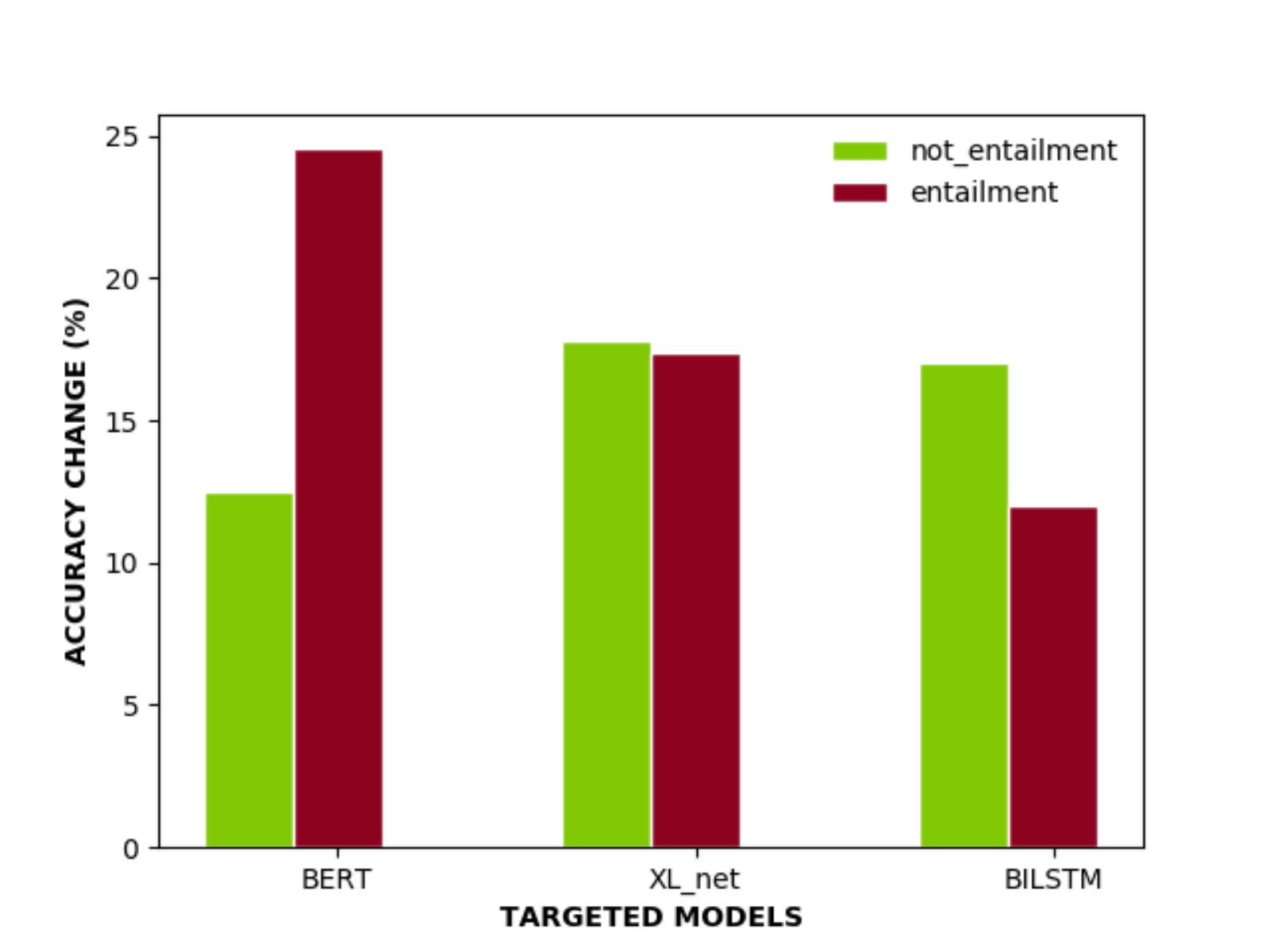}
			\caption{QNLI 3 Word}
		\end{subfigure}%
		\begin{subfigure}{.5\textwidth}
			\centering
			\includegraphics[width=60mm, height=40mm]{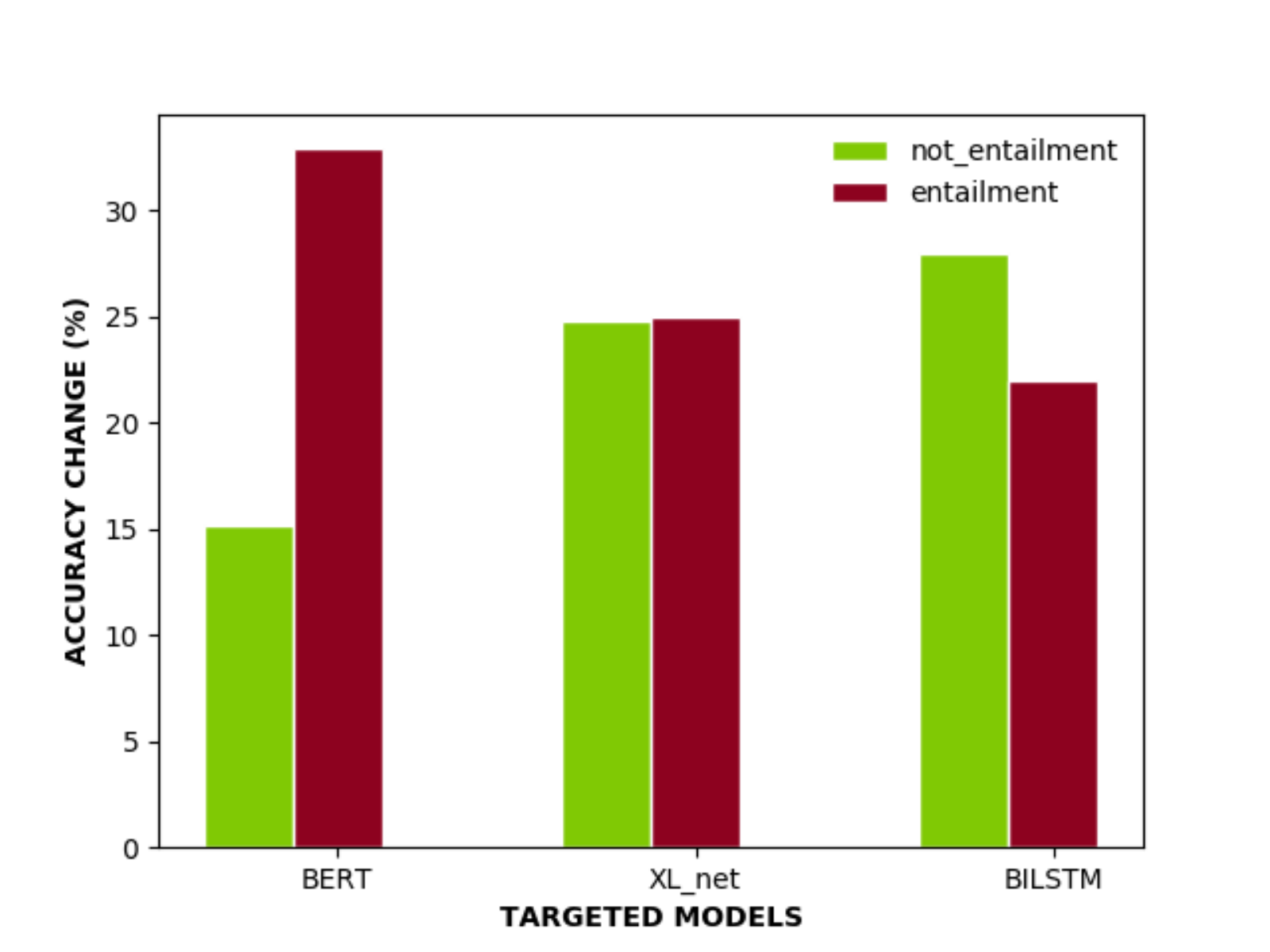}
			\caption{QNLI 4 Word}
		\end{subfigure}%
		\caption{Classwise accuracy change for QNLI on the three target models with maximum perturbation allowed upto 3(left subfigure) and 4(right subfigure) words.}
		\label{fig:Class_acc_diff_qnli}
\end{figure*}

Figures \ref{fig:Class_acc_diff_3_word}, \ref{fig:Class_acc_diff_4_word} and \ref{fig:Class_acc_diff_qnli} show the classwise accuracy changes/drops for various datasets. For \textit{MNLI}, the \% drop in accuracy for \textit{entailment} class is particularly high. This is primarily due to high lexical overlap between the premise and hypothesis 
, thus changing the lexical structure of hypothesis leads the models to false predictions of \emph{entailment} class. \textit{SST-2} being a balanced dataset shows similar performance degradation for both the classes. For \textit{QQP}, the class with label 1(\emph{Similar}) has much higher \% drop accuracy. We hypothesize the primary reason for this to be the \textit{skewness} of the dataset for class with label 0(\emph{Dissimilar}) due to which it is easier to fool the models by exploiting this bias of training and easily flip the model predictions from class 1 to class 0.

\newpage

\begin{table*}[ht]
\tiny

\centering
\setlength\tabcolsep{3.5pt}
\begin{tabular}{lcccccccccccccc}
\\
\hline
\textbf{}   &  \multicolumn{5}{c}{\bf \emph{BERT}}  &  \multicolumn{5}{c}{\bf \emph{XLNeT}} & \multicolumn{4}{c}{\bf \emph{BiLSTM + Attn + ELMo}} \\
\cline{2-5}   \cline{7-10} \cline{12-15}
&  SST-2  & MNLI(m / mm) & QQP  & QNLI &  &  SST-2  & MNLI(m / mm) & QQP  & QNLI &  &  SST-2  & MNLI(m / mm) & QQP  & QNLI \\

\hline
\emph{Pre Attack Acc.} & 92.14 & 82.7 / 84.88 & 90.5 & 84.65 &  & 93.3 & 85.00 / 85.38 & 88.93 & 83.21 &  & 88.68 & 70.1 / 71.36 & 81.4 & 76.2 \\                                                      

&\\

\hline
\textbf{} &  \multicolumn{14}{c}{\bf \emph{Method: TextFooler}} \\
\hline
\emph{Avg. Accuracy Drops} & \underline{1.61} & \underline{2.54} / \underline{2.83} & 1.95 & \underline{2.13} & & \underline{1.80} & \underline{3.04} / \underline{2.76} & 2.53 & \underline{2.48} &  & \underline{2.86} & \underline{3.92} / \underline{3.53} & 2.49 & 1.72 \\
\emph{Max. Accuracy Drops} & 12.32 & 35.61 / 37.89 \textbf{**} & 25.71 & 9.07 &  & 13.58 & 38.93 / 37.81 & 27.36\textbf{**} & 6.83 &  & 15.72 & 35.91 / 35.63\textbf{**} & 26.39 & 10.72\textbf{**} \\
\emph{Semantic Sim} & 0.76 & 0.73 / 0.71 & 0.71 & 0.82 & & 0.77 & 0.65 / 0.67 & 0.73 & 0.80 &  & 0.74 & 0.69 / 0.71 & 0.72 & 0.80 \\

&\\

\hline           
\textbf{} &  \multicolumn{14}{c}{\bf \emph{Method: Genetic}} \\
\hline
\emph{Avg. Accuracy Drops} & 1.45 & 2.27 / 2.62 & \underline{1.99} & 1.86 &  & 1.67 & 2.76  / 2.53 & \underline{2.78} & 2.30 &  & 2.49 & 3.51 / 3.40 & \underline{2.68} & \underline{1.74} \\
\emph{Max. Accuracy Drops} & 13.52\textbf{**} & 36.58 / 37.63 & 26.03\textbf{**} & 10.96\textbf{**} &  & 15.98\textbf{**} & 41.26 / 42.34 \textbf{**} & 26.52 & 8.09\textbf{**} &  & 16.34\textbf{**} & 33.91 / 32.38 & 27.93\textbf{**} & 8.37 \\
\emph{Semantic Sim} & 0.74 & 0.68 / 0.65 & 0.70 & 0.82 & & 0.74 & 0.62 / 0.63 & 0.68 & 0.82 &  & 0.71 & 0.60 / 0.59 & 0.69 & 0.81 \\
&\\

\hline
\textbf{} &  \multicolumn{14}{c}{\bf \emph{Method: Ours}} \\
\hline
\emph{Avg. Accuracy Drops} & \textbf{7.82} & \textbf{26.07} / \textbf{25.88} & \textbf{8.03} & \textbf{6.52} &  & \textbf{7.59} & \textbf{25.48} / \textbf{26.13} & \textbf{7.95} & \textbf{6.39} &  & \textbf{9.32} & \textbf{28.43} / \textbf{28.92} & \textbf{8.27} & \textbf{6.76} \\
\emph{Max. Accuracy Drops} & 33.67\textbf{*} & 66.24 / 66.49\textbf{*} & 33.70\textbf{*} & 17.89\textbf{*} &  & 34.17\textbf{*} & 65.84 / 66.91 \textbf{*} & 35.49 \textbf{*} & 18.19 \textbf{*} &  & 29.39\textbf{*} & 49.17 / 48.81 \textbf{*} & 29.27\textbf{*} & 18.41 \textbf{*} \\
\emph{Semantic Sim} & 0.78 & 0.72 / 0.70 & 0.73 & 0.81 &  & 0.80 & 0.69 / 0.68 & 0.71 & 0.79 &  & 0.77 & 0.72 / 0.73 & 0.74 & 0.79  \\
\hline

\end{tabular}
\caption{Word adversarial attack results by perturbing atmost \textbf{4} words. The table shows accuracy drops for various methods as well the average semantic score of the adversarial sentences. \emph{Pre Attack Acc} represents the accuracy on original sentences. \emph{Avg. Accuracy Drops} represent the drops in accuracy averaged over the adversarial candidates, whereas \emph{Max. Accuracy Drops} represent the maximum drop in accuracy achieved if any adversarial candidate in the set of $K=20$ succeeds. Higher the drops, more successfull the attack. Here, \emph{m} and \emph{mm} represent the \emph{matched} and \emph{mismatched} versions of the dev set respectively. For Average Drops, the best results are marked in \textbf{bold} whereas the second best \underline{underlined}. For Maximum Drops, the best are marked by \textbf{*}, whereas the second best by \textbf{**}.} 
\label{table:word_results_4_perturb}
\end{table*}

\begin{table*}[ht]
\tiny

\centering
\setlength\tabcolsep{3.5pt}
\begin{tabular}{lcccccccccccccc}
\\
\hline
\textbf{}   &  \multicolumn{5}{c}{\bf \emph{BERT}}  &  \multicolumn{5}{c}{\bf \emph{XLNeT}} & \multicolumn{4}{c}{\bf \emph{BiLSTM + Attn + ELMo}} \\
\cline{2-5}   \cline{7-10} \cline{12-15}
&  SST-2  & MNLI(m / mm) & QQP  & QNLI &  &  SST-2  & MNLI(m / mm) & QQP  & QNLI &  &  SST-2  & MNLI(m / mm) & QQP  & QNLI \\

\hline
\emph{Pre Attack Acc.} & 92.14 & 82.7 / 84.88 & 90.5 & 84.65 &  & 93.3 & 85.00 / 85.38 & 88.93 & 83.21 &  & 88.68 & 70.1 / 71.36 & 81.4 & 76.2 \\                                                      

&\\

\hline
\textbf{} &  \multicolumn{14}{c}{\bf \emph{Method: TextFooler}} \\
\hline
\emph{Avg. Accuracy Drops} & \underline{2.02} & \underline{3.32} / \underline{3.28} & 2.81 & \underline{3.04} & & \underline{2.68} & \underline{3.83} / 3.50 & 2.98 & \underline{3.83} &  & \underline{3.64} & 4.93 / 4.76 & 3.08 & 2.66 \\
\emph{Max. Accuracy Drops} & 15.32 & 42.63 / 43.91 & 28.92 & 13.53\textbf{**} &  & 19.32 & 47.23 / 45.98 & 32.82\textbf{**} & 9.20\textbf{**} &  & 17.36 & 39.86 / 41.07 \textbf{**} & 29.63 & 14.35\textbf{**} \\
\emph{Semantic Sim} & 0.72 & 0.65 / 0.63 & 0.69 & 0.73 & & 0.70 & 0.60 / 0.59 & 0.68 & 0.75 &  & 0.71 & 0.68 / 0.67 & 0.67 & 0.74 \\

&\\

\hline           
\textbf{} &  \multicolumn{14}{c}{\bf \emph{Method: Genetic}} \\
\hline
\emph{Avg. Accuracy Drops} & 1.98 & 3.11 / 3.16 & \underline{3.07} & 3.00 &  & 2.53 & 3.52 / \underline{3.63} & \underline{3.34} & 3.69 &  & 3.61 & \underline{4.96} / \underline{4.83} & \underline{3.26} & \underline{2.83} \\
\emph{Max. Accuracy Drops} & 17.89\textbf{**} & 45.56 / 47.09 \textbf{**} & 30.81\textbf{**} & 13.36 &  & 23.72\textbf{**} & 50.72 / 50.82\textbf{**} & 30.80 & 8.99 &  & 18.83\textbf{**} & 37.04 / 38.45 & 31.05\textbf{**} & 13.68 \\
\emph{Semantic Sim} & 0.68 & 0.62 / 0.64 & 0.66 & 0.74 & & 0.68 & 0.58 / 0.57 & 0.65 & 0.76 &  & 0.69 & 0.57 / 0.55 & 0.68 & 0.75 \\
&\\

\hline
\textbf{} &  \multicolumn{14}{c}{\bf \emph{Method: Ours}} \\
\hline
\emph{Avg. Accuracy Drops} & \textbf{8.26} & \textbf{28.91} / \textbf{28.75} & \textbf{9.47} & \textbf{8.04} &  & \textbf{8.57} & \textbf{27.62} / \textbf{28.28} & \textbf{9.28} & \textbf{7.93} &  & \textbf{11.21} & \textbf{31.29} / \textbf{31.83} & \textbf{9.84} & \textbf{8.37} \\
\emph{Max. Accuracy Drops} & 38.74\textbf{*} & 70.26 / 69.97 \textbf{*} & 36.18\textbf{*} & 22.26\textbf{*} &  & 37.91\textbf{*} & 71.04 / 70.84 \textbf{*} & 37.82\textbf{*} & 23.11\textbf{*} &  & 34.82\textbf{*} & 54.17 / 55.08 \textbf{*} & 32.08\textbf{*} & 21.59\textbf{*} \\
\emph{Semantic Sim} & 0.73 & 0.68 / 0.66 & 0.67 & 0.75 &  & 0.74 & 0.62 / 0.61 & 0.66 & 0.72 &  & 0.72 & 0.69 / 0.68 & 0.70 & 0.73  \\
\hline

\end{tabular}
\caption{Word adversarial attack results by perturbing atmost \textbf{5} words. The table shows accuracy drops for various methods as well the average semantic score of the adversarial sentences. \emph{Pre Attack Acc} represents the accuracy on original sentences. \emph{Avg. Accuracy Drops} represent the drops in accuracy averaged over the adversarial candidates, whereas \emph{Max. Accuracy Drops} represent the maximum drop in accuracy achieved if any adversarial candidate in the set of $K=20$ succeeds. Higher the drops, more successfull the attack. Here, \emph{m} and \emph{mm} represent the \emph{matched} and \emph{mismatched} versions of the dev set respectively. For Average Drops, the best results are marked in \textbf{bold} whereas the second best \underline{underlined}. For Maximum Drops, the best are marked by \textbf{*}, whereas the second best by \textbf{**}.} 
\label{table:word_results_5_perturb}
\end{table*}

\begin{table*}[ht]
\tiny

\centering
\setlength\tabcolsep{3.5pt}
\begin{tabular}{lccccccccccc}
\\
\hline
\textbf{}   &  \multicolumn{4}{c}{\bf \emph{BERT}}  &  \multicolumn{4}{c}{\bf \emph{XLNeT}} & \multicolumn{3}{c}{\bf \emph{BiLSTM + Attn + ELMo}} \\
\cline{2-4}   \cline{6-8} \cline{10-12}
& 3 Word  & 4 Word & 5 Word &  & 3 Word  & 4 Word & 5 Word &  & 3 Word  & 4 Word & 5 Word \\

\hline
\emph{Pre Attack MSE} & 0.81 & 0.81 & 0.81 &  & 1.21 & 1.21 & 1.21 &  & 1.34 & 1.34 & 1.34 \\                                                      

&\\

\hline
\textbf{} &  \multicolumn{11}{c}{\bf \emph{Method: TextFooler}} \\
\hline
\emph{Post Attack MSE} & \underline{1.93} & 2.17 & 2.34 & & \underline{1.67} & \underline{1.82} & \underline{2.06} & & 1.83 & 2.09 & 2.43 \\
\emph{Semantic Sim} & 0.71 & 0.68 & 0.60 & & 0.68 & 0.66 & 0.59 & & 0.72 & 0.67 & 0.63 \\
&\\

\hline
\textbf{} &  \multicolumn{11}{c}{\bf \emph{Method: Genetic}} \\
\hline
\emph{Post Attack MSE} & 1.89 & \underline{2.23} & \textbf{2.41} & & 1.53 & 1.76 & 2.04 & & \underline{1.94} & \underline{2.16} & \underline{2.54} \\
\emph{Semantic Sim} & 0.67 & 0.64 & 0.58 & & 0.64 & 0.62 & 0.57 & & 0.70 & 0.66 & 0.62 \\
&\\

\hline
\textbf{} &  \multicolumn{11}{c}{\bf \emph{Method: Our}} \\
\hline
\emph{Post Attack MSE} & \textbf{1.97} & \textbf{2.24} & \underline{2.38} & & \textbf{1.78} & \textbf{1.99} & \textbf{2.07} & & \textbf{2.02} & \textbf{2.35} & \textbf{2.58} \\
\emph{Semantic Sim} & 0.72 & 0.67 & 0.61 & & 0.70 & 0.65 & 0.60 & & 0.74 & 0.68 & 0.63 \\
&\\

\end{tabular}
\caption{Adversarial attack results of all methods on the regression dataset: \emph{STS-B} for various word perturbation levels: \textbf{3,4} and \textbf{5}. \emph{Pre Attack MSE} represents the \emph{mean squared error} on original sentences. \emph{Post Attack MSE} represents the mean squared error averaged over the adversarial candidates. \emph{Semantic Sim} represents the average semantic similarity of the adversaries to the original sentences. Higher the value of \emph{Post Attack MSE}, more successfull the attack. Best results are marked in \textbf{bold}, while the second best are \underline{underlined}.} 
\label{table:sts-b_results}
\end{table*}

\section{Dataset Descriptions}
\noindent\textbf{SST-2}: is a binary classification dataset of reviews from movies annotated by humans. 

\noindent\textbf{MNLI}: is a huge collection of sentence pairs termed as \emph{hypothesis} and \emph{premise} with their textual entailment annotations. Given a premise and a hypothesis sentence, the task is to predict whether the premise entails, contradicts or shows neutrality to the hypothesis. The dev portion of this dataset contains two subcomponents marked - \emph{matched(m)} and \emph{mismatched(mm)}. We evaluate our method on both subcomponents separately.

\noindent\textbf{QNLI}: cast question-paragraph pairs from SQuAD into a classification task forming a pair between each question and each sentence in the corresponding context (from the paragraph) with high lexical overlap. The task at hand is to determine whether the context sentence contains the answer to the question.

\noindent\textbf{QQP}: is a huge collection of questions from the community question-answering website \emph{Quora}. The task is to determine whether two questions are semantically equivalent marked by $1$ (\emph{semantically similar}) or $0$ (\emph{semantically dissimilar}).

\noindent\textbf{STS-B}: is a collection of sentence pairs drawn from various sources. The task is to determine the semantic score between two sentences ranging from $1$ to $5$, thus marked as a regression task.

\section{Attacked Models}
We provide a brief description of the attacked models here. For both \emph{BERT} and \emph{XLNet}, we use the PyTorch implementation\footnote{\footnotesize \url{https://github.com/huggingface/transformers}}, with $12$ hidden layers, $768$ hidden units, $12$ attention heads, and sequence lengths truncated to $128$. For \emph{BiLSTM model with attention}, we use a $2$ layer bidirectional LSTM with hidden dimension embeddings of size $1500$ as well as input embeddings initialized using ELMo embeddings, having a MLP classifier with $512$ hidden units. We use the implementation available here \footnote{\footnotesize \url{https://github.com/nyu-mll/GLUE-baselines}}.

\begin{table*}[ht]
        \small
		
		\centering
		\begin{tabular}{p{2cm}p{10cm}p{2.5cm}}
		 \\
        \hline
        \textbf{Sent Type}  & \multicolumn{1}{c}{\bf Input Sentence} & \textbf{Model Prediction} \\
        \hline
        &\\
        \multicolumn{3}{c}{\bf \emph{Model: BiLSTM, Task: SST-2}}\\
        \hline
        Original & Allow us to hope that Nolan is poised to embark on a major career as a commercial \textcolor{blue}{yet inventive filmmaker}. & \textbf{Positive} \\
        Adversarial & Allows us to hope that Nolan is poised to embark on a major career as a commercial \textcolor{red}{but creative director}. & \textbf{Negative} \\
        \hline
        
        &\\
        \multicolumn{3}{c}{\bf \emph{Model: XLNeT, Task: SST-2}}\\
        \hline
        Original & In its best moments, \textcolor{blue}{resembles} a bad high school production of grease, without benefit of \textcolor{blue}{song}. & \textbf{Negative} \\
        Adversarial & In its best moments, \textcolor{red}{remembering} a bad high school production of grease, without benefit of \textcolor{red}{anthems}. & \textbf{Positive} \\
        \hline
        
        &\\
        \multicolumn{3}{c}{\bf \emph{Model: BiLSTM, Task: MNLI}}\\
        \hline
        Premise & There are no shares of a stock that might someday come back, just piles of options as worthless as those shares of Cook's american business alliance . &  \\
        Original & Cook's \textcolor{blue}{american} business alliance caused shares of stock to come \textcolor{blue}{back}. & \textbf{Contradiction} \\
        Adversarial & Cook's \textcolor{red}{latino} business alliance caused shares of stock to come \textcolor{red}{backwards}.  & \textbf{Entailment} \\
        \hline
        
        &\\
        \multicolumn{3}{c}{\bf \emph{Model: XLNeT, Task: MNLI}}\\
        \hline
        Premise & If that investor were willing to pay extra for the security of limited downside, she could buy put options with a strike price of \$98, which would lock in her profit on the shares at \$18, less whatever the options cost. &  \\
        Original & The \textcolor{blue}{strike} price could be \$8. & \textbf{Contradiction} \\
        Adversarial & The \textcolor{red}{shelling} price could be \$8. & \textbf{Neutral} \\
        \hline
        
		\end{tabular}
		\caption{Samples from the target model agnostic adversaries generated by our method. The last column shows the predictions of the target models (over the original sentences which were predicted correctly).}
		\label{table:adversarial_samples}
\end{table*}

\begin{table*}[ht]
        \small
		
		\centering
		\begin{tabular}{p{2cm}p{10cm}p{2.5cm}}
		 \\
        \hline
        \textbf{Sent Type}  & \multicolumn{1}{c}{\bf Input Sentence} & \textbf{Model Prediction} \\
        \hline
        &\\
        \multicolumn{3}{c}{\bf \emph{Model: BiLSTM, Task: QQP}}\\
        \hline
        First Sent & What people who you've never met have influenced your life the most? & \\
        Original & Who are the \textcolor{blue}{people} you have never \textcolor{blue}{met} who have had the greatest influence on your \textcolor{blue}{life} ? & \textbf{Similar} \\
        Adversarial & Who are the \textcolor{red}{nationals} you have never \textcolor{red}{encountered} who have had the greatest influence on your \textcolor{red}{vida} ? & \textbf{Dissimilar} \\
        \hline
        
        &\\
        \multicolumn{3}{c}{\bf \emph{Model: XLNeT, Task: QQP}}\\
        \hline
        First Sent & How do you get better grades ? & \\
        Original & How can I \textcolor{blue}{dramatically} \textcolor{blue}{improve} my grades ? & \textbf{Similar} \\
        Adversarial & How can I \textcolor{red}{immensely} \textcolor{red}{boost} my grades ? & \textbf{Dissimilar} \\
        \hline
        
        &\\
        \multicolumn{3}{c}{\bf \emph{Model: BiLSTM, Task: QNLI}}\\
        \hline
        Premise &  What is another possible explanation for the source of the signals ? &  \\
        Original & He expanded on the signals he heard in a 9 february 1901 Collier's weekly article talking with planets where he said it had not been immediately \textcolor{blue}{apparent} to him that he was hearing intelligently controlled \textcolor{blue}{signals} and that the signals could come from \textcolor{blue}{mars}, venus or other planets.  & \textbf{Not Entailment} \\
        Adversarial & He expanded on the signals he heard in a 9 february 1901 Collier's weekly article talking with planets where he said it had not been immediately \textcolor{red}{noticeable} to him that he was hearing intelligently controlled \textcolor{red}{gesture} and that the signals could come from \textcolor{red}{mar}, venus or other planets.  & \textbf{Entailment} \\
        \hline
        
        &\\
        \multicolumn{3}{c}{\bf \emph{Model: XLNeT, Task: QNLI}}\\
        \hline
        Premise & What religion did tesla grow up in ?  &  \\
        Original & Later in his life, he did not consider himself to be a \textcolor{blue}{believer} in the orthodox \textcolor{blue}{sense}, and opposed religious \textcolor{blue}{fanaticism}.  & \textbf{Not Entailment} \\
        Adversarial & Later in his life, he did not consider himself to be a \textcolor{red}{devotee} in the orthodox \textcolor{red}{vein}, and opposed religious \textcolor{red}{homophobic}.  & \textbf{Entailment} \\
        \hline
        
		\end{tabular}
		\caption{Examples from the target model agnostic adversaries generated by our method. The last column shows the predictions of the target models (over the original setences which were predicted correctly).}
		\label{table:1_adversarial_samples1}
		
\end{table*}

\begin{table*}[ht]
        \normalsize
		\centering
		\begin{tabular}{p{15cm}}
		\\
        \hline
        
        \multicolumn{1}{c}{\bf \emph{SST-2}} \\
        \hline
        All that's \textcolor{blue}{missing} is the \textcolor{blue}{spontaneity} originality and \textcolor{blue}{delight}. \\
        If \textcolor{blue}{Steven} Soderbergh's \textcolor{blue}{solaris} is a \textcolor{blue}{failure}, it is a glorious failure.\\
        \hline
        
        \multicolumn{1}{c}{\bf \emph{QQP}} \\
        \hline
        What are the requirements to become president in the united \textcolor{blue}{states} and how are they requirements \textcolor{blue}{different} in \textcolor{blue}{France}? \\
        What are the top books an aspiring \textcolor{blue}{teen} \textcolor{blue}{entrepreneur} should \textcolor{blue}{read}? \\
        \hline

        \multicolumn{1}{c}{\bf \emph{STS-B}} \\
        \hline
        The \textcolor{blue}{dog} is \textcolor{blue}{playing} with a \textcolor{blue}{plastic} container. \\
        Someone is \textcolor{blue}{drilling} a \textcolor{blue}{hole} in a piece of \textcolor{blue}{wood}. \\
        \hline
        
        \multicolumn{1}{c}{\bf \emph{QNLI}} \\
        \hline
        The Lazienki \textcolor{blue}{park} \textcolor{blue}{covers} the \textcolor{blue}{area} of 76 Ha. \\
        The \textcolor{blue}{building} was \textcolor{blue}{designed} by architects \textcolor{blue}{Marek} Budzyński and Zbigniew and opened on 15 December 1999.\\
        \hline
        
        \multicolumn{1}{c}{\bf \emph{MNLI}} \\
        \hline
        You don't \textcolor{blue}{want} to \textcolor{blue}{push} the \textcolor{blue}{button} lightly, but rather punch it \textcolor{blue}{hard}. \\
        The slopes between the Vosges and \textcolor{blue}{Rhine} valley are the only place \textcolor{blue}{appropriate} for \textcolor{blue}{vineyards}. \\
        \hline
        
		\end{tabular}
		\caption{Top $M$($= 3$) words selected as important by our enhanced ON-LSTM method to be replaced marked in blue.}
		\label{table:important_words}
		
\end{table*}

\begin{table*}[ht]
        \small

\centering
\begin{tabular}{p{2cm}p{10cm}p{2.5cm}}
\\
        \hline
        \textbf{Sent Type}  & \multicolumn{1}{c}{\bf Input Sentence} & \textbf{Model Prediction} \\
        \hline
       
        &\\
        \multicolumn{3}{c}{\bf \emph{Model: XLNeT, Task: SST-2}}\\
        \hline
        Original & An absurdist comedy about alienation, se\textcolor{blue}{pa}ation and loss.  & \textbf{Negative} \\
        Adversarial & An absurdist comedy about alienation, se\textcolor{red}{Zj}ration and loss. & \textbf{Positive} \\
        \hline
       
        &\\
        \multicolumn{3}{c}{\bf \emph{Model: BERT, Task: SST-2}}\\
        \hline
        Original & A subject like this should inspire reac\textcolor{blue}{ti}on in its audience the pianist does not. & \textbf{Negative} \\
        Adversarial & A subject like this should inspire reac\textcolor{red}{Yj}on in its audience the pianist does not. & \textbf{Positive} \\
        \hline

        &\\
        \multicolumn{3}{c}{\bf \emph{Model: XLNeT, Task: QQP}}\\
        \hline
        First Sent & Online gaming with irl friends is more fun. Why do you play with randoms? & \\
        Original & Playing dota2 with irl fr\textcolor{blue}{i}e\textcolor{blue}{n}ds is more fun. Why do you play with randoms? & \textbf{Similar} \\
        Adversarial & Playing dota2 with irl fr\textcolor{red}{.}e\textcolor{red}{*}ds is more fun. Why do you play with randoms? & \textbf{Dissimilar} \\
        \hline
       
        &\\
        \multicolumn{3}{c}{\bf \emph{Model: BERT, Task: QQP}}\\
        \hline
        First Sent & What are some of the best jokes you've ever heard? & \\
        Original & What is the funniest joke you ever he\textcolor{blue}{ar}d? & \textbf{Dissimilar} \\
        Adversarial & what is the funniest joke you ever he\textcolor{red}{x9}d?  & \textbf{Similar} \\
        \hline
       
        &\\
        \multicolumn{3}{c}{\bf \emph{Model: XLNeT, Task: QNLI}}\\
        \hline
        Premise & Who reportedly wanted tesla's company? & \\
        Original & There have been numerous accounts of women vying for tesla's a\textcolor{blue}{f}fec\textcolor{blue}{t}ion, even some madly in love with him.  & \textbf{Not Entailment} \\
        Adversarial & There have been numerous accounts of women vying for tesla's a\textcolor{red}{I}fec\textcolor{red}{c}ion, even some madly in love with him.  & \textbf{Entailment} \\
        \hline
       
        &\\
        \multicolumn{3}{c}{\bf \emph{Model: BERT, Task: QNLI}}\\
        \hline
        Premise & Who served his dinner? & \\
        Original & He dined alone, except on the rare occasions when he would give a dinner to a group to meet his social obl\textcolor{blue}{i}ga\textcolor{blue}{t}ions. & \textbf{Not Entailment} \\
        Adversarial & He dined alone, except on the rare occasions when he would give a dinner to a group to meet his social obl\textcolor{red}{k}ga\textcolor{red}{,}ions.   & \textbf{Entailment} \\
        \hline

        &\\
        \multicolumn{3}{c}{\bf \emph{Model: XLNeT, Task: MNLI}}\\
        \hline
        Premise & Look out for that overseer up there. & \\
        Original & Watch out that you do n\textcolor{blue}{o}t bump your head on the overseer. & \textbf{Neutral} \\
        Adversarial & Watch out that you do N\textcolor{red}{7}t bump your head on the overseer. & \textbf{Entailment} \\
        \hline

        &\\
        \multicolumn{3}{c}{\bf \emph{Model: BERT, Task: MNLI}}\\
        \hline
        Premise & A re-created street of colonial Macau is lined with traditional chinese shops. & \\
        Original & You'll find plenty of au\textcolor{blue}{t}he\textcolor{blue}{n}tic, old world restaurants on that street . & \textbf{Neutral} \\
        Adversarial & You'll find plenty of au\textcolor{red}{H}he\textcolor{red}{'}tic, old world restaurants on that street. & \textbf{Contradiction} \\
        \hline

        &\\
        \multicolumn{3}{c}{\bf \emph{Model: XLNeT, Task: STS-B}}\\
        \hline
        First Sent & It is possible, but it will have to be a docile female betta and a bigish tank.  & \\
        Original & We tr\textcolor{blue}{ie}d putting a male betta in a community tank once. & \textbf{2.4} \\
        Adversarial & We tr\textcolor{red}{y.}d putting a male betta in a community tank once.  & \textbf{3.45} \\
        \hline
       
        &\\
        \multicolumn{3}{c}{\bf \emph{Model: BERT, Task: STS-B}}\\
        \hline
        First Sent & One thing you seem to be forgetting regarding myths, is they are extremely prevalent stories.   & \\
        Original &  I noticed you said movie critics \textcolor{blue}{en}joy mythological references in a film, but do audiences? & \textbf{1.2} \\
        Adversarial & I noticed you said movie critics \textcolor{red}{2O}joy mythological references in a film, but do audiences?      & \textbf{0.71} \\
        \hline
       
\end{tabular}
\caption{Examples from the target model agnostic adversaries over characters  generated by our method. The last column shows the predictions of the target models (over the original setences which were predicted correctly).}
\label{table:2_adversarial_samples2}

\end{table*}

\begin{table*}[ht]
        \small
		
		\centering
		\begin{tabular}{p{2cm}p{10cm}p{2.5cm}}
		 \\
		 \multicolumn{3}{c}{\bf Model: BiLSTM}\\
		 \vspace*{0.1cm} \\
        \hline
        \textbf{Sent Type}  & \multicolumn{1}{c}{\bf Input Sentence} & \textbf{Model Prediction} \\
        \hline
        
         &\\
        \multicolumn{3}{c}{\bf \emph{Task: SST-2}}\\
        \hline
        Original & Aside from minor tinkering, this is the same movie you probably loved in 1994, except that it lo\textcolor{blue}{ok}s even better & \textbf{Positive} \\
        Adversarial & Aside from minor tinkering, this is the same movie you probably loved in 1994, except that it lo\textcolor{red}{XE}s even better . & \textbf{Negative} \\
        \hline
        
        &\\
        \multicolumn{3}{c}{\bf \emph{Task: QQP}}\\
        \hline
        First Sent & What are some mind blowing car technology gadgets that exist in 2016 that most people don't know about? & \\
        Original & What are the most advanced car gadgets that people don't know about y\textcolor{blue}{et}? & \textbf{Similar} \\
        Adversarial & What are the most advanced car gadgets that people don't know about y\textcolor{red}{Ft} ? & \textbf{Dissimilar} \\
        \hline
        
        &\\
        \multicolumn{3}{c}{\bf \emph{Task: QNLI}}\\
        \hline
        Premise & Who said Tesla had a distinguished sweetness? & \\
        Original & His loyal secretary, Dorothy Skerrit, wrote his genial smile and nobility of bearing always denoted the gentlemanly charac\textcolor{blue}{te}ristics that were so ingrained in his soul. & \textbf{Entailment} \\
        Adversarial & His loyal secretary, Dorothy Skerrit, wrote his genial smile and nobility of bearing always denoted the gentlemanly charac\textcolor{red}{qC}ristics that were so ingrained in his soul.  & \textbf{Not Entailment} \\
        \hline
        
        &\\
        \multicolumn{3}{c}{\bf \emph{Task: MNLI}}\\
        \hline
        Premise & Yeah it's true it is in in fact I have a friend of mine that moved to North Carolina she's um an emergency room nurse she does the operating room.  & \\
        Original & This person I'm close to is an eme\textcolor{blue}{r}ge\textcolor{blue}{n}cy room nurse at a hospital in North Carolina  & \textbf{Entailment} \\
        Adversarial & This person I'm close to is an eme\textcolor{red}{e}ge\textcolor{red}{K}cy room nurse at a hospital in North Carolina. & \textbf{Neutral} \\
        \hline
        
        &\\
        \multicolumn{3}{c}{\bf \emph{Task: STS-B}}\\
        \hline
        First Sent & Keep in mind that you can easily swear without swearing.  & \\
        Original & I th\textcolor{blue}{in}k stephen king's comments are helpful in this regard. & \textbf{1.20} \\
        Adversarial & I th\textcolor{red}{M"}k stephen king's comments are helpful in this regard. & \textbf{2.36} \\
        \hline

        \end{tabular}
		\caption{Examples from the target model agnostic adversaries over characters  generated by our method. The last column shows the predictions of the target model (over the original setences which were predicted correctly).}
		\label{table:3_adversarial_samples3}
		
\end{table*}

\end{document}